\documentclass[journal]{IEEEtran}
\ifCLASSINFOpdf
   \usepackage[pdftex]{graphicx}
\else
\fi
\usepackage{amsmath}
\usepackage[linesnumbered, ruled, vlined, english]{algorithm2e}

\usepackage{amssymb}
\usepackage{cite}
\usepackage[font=footnotesize]{caption, subcaption}
\usepackage{xcolor}
\usepackage{bm}
\usepackage{multirow}
\usepackage{float}

\SetCommentSty{mycommfont}

\begin{document}
%
\title{Dual-Arm In-Hand Manipulation and Regrasping Using Dexterous Manipulation Graphs}
%
%
%

\author{Silvia~Cruciani,
		Kaiyu~Hang, Christian~Smith and~Danica~Kragic
\thanks{S. Cruciani, C. Smith and D. Kragic are with the Division of Robotics, Perception and Learning, EECS, KTH Royal Institute of Technology, Stockholm, Sweden. e-mail: cruciani@kth.se, ccs@kth.se, dani@kth.se }
\thanks{K. Hang is with the Department of Mechanical Engineering and Material Science, Yale University, New Haven, CT USA. e-mail: kaiyu.hang@yale.edu}
\thanks{This work was supported by the Swedish Foundation for Strategic Research project GMT14-0082 FACT.}%
}

%
%

\markboth{ 
	}%
{
	}
%



\maketitle

\begin{abstract}
This work focuses on the problem of in-hand manipulation and regrasping of objects with parallel grippers. We propose Dexterous Manipulation Graph (DMG) as a representation on which we define planning for in-hand manipulation and regrasping. The DMG is a disconnected undirected graph that represents the possible motions of a finger along the object's surface. We formulate the in-hand manipulation and regrasping problem as a graph search problem from the initial to the final configuration. The resulting plan is a sequence of coordinated in-hand pushing and regrasping movements. We propose a dual-arm system for the execution of the sequence where both hands are used interchangeably. We demonstrate our approach on an ABB Yumi robot tasked with different grasp reconfigurations.
\end{abstract}


%
\IEEEpeerreviewmaketitle

\section{Introduction}
Motions such as sliding, rotating and regrasping the object are natural to humans and are enabled by the dexterous structure of the human hand. For robots, it is still challenging to execute such manipulations that go beyond simple pick-and-place. One reason is that artificial hands that resemble human dexterity are not yet widespread in robotics due to inherent difficulties in design and control. Still, the most common end-effector of a robot arm is a simple parallel gripper. The problem of correctly grasping an object is of fundamental importance for object and tool use, both in terms of industrial and service robots applications. 

In this work, we focus on achieving in-hand manipulation and regrasping of rigid objects using parallel grippers.
The main challenge in manipulating objects with a parallel gripper lies in the lack of degrees of freedom (DOF), as these grippers can only open and close. However, parallel grippers are not just a tool for grasping, but they can be used for object pushing, in line with the concept of \emph{extrinsic dexterity}. With extrinsic dexterity, the lack of DOF can be compensated using external supports, such as contact surfaces, gravity and inertial forces.

In our work, we exploit a dual-arm system, where the inability of the gripper is compensated by the two arms.
This is naturally built upon the fact that dual-arm robots are becoming more common (e.g. Rethink Robotics Baxter, ABB Yumi) and we believe that their exploitation will help overcome various challenges that relate to advanced object interaction.

\begin{figure}
	\centering
	\includegraphics[width=0.42\textwidth]{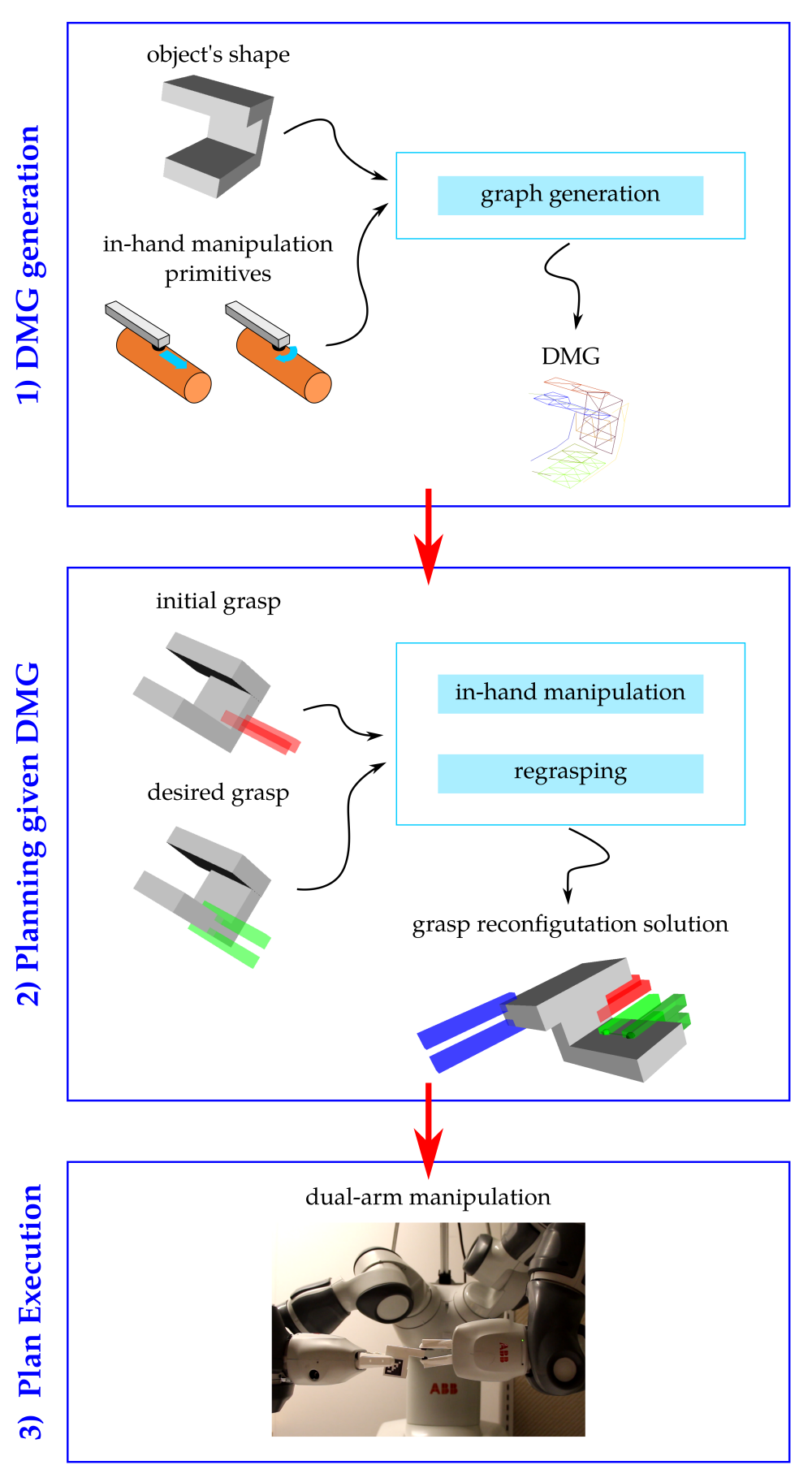}
	\caption{An overview of our system. 1) We generate the DMG from the object's shape. 2) With the DMG, we plan a solution to from an initial configuration (red) to the desired one (green). 3) The plan is executed by a dual-arm robot.}\label{fig_overview}
\end{figure}

Our particular focus in this paper is on in-hand object manipulation: how to move an object from an initial grasp configuration to a desired one. 
We formulate the in-hand manipulation and regrasping problem as a planning problem from the initial to the final configuration, using Dexterous Manipulation Graphs (DMG). 
Our system consists of three main steps:
\begin{enumerate}
	\item DMG generation, described in section~\ref{sec_graph_generation}.
	\item Planning given DMG, described in sections~\ref{sec_object_manipulation_planning} and~\ref{sec_regrasp_planning}.
	\item Plan execution, described in section~\ref{sec_execution}.
\end{enumerate}
These steps are shown in Fig.~\ref{fig_overview}. The starting point is the DMG generation from the object's shape. The DMG represents the possible motions of a finger on the object's surface. Therefore, it depends on the kind of motions considered, which we define as in-hand manipulation primitives. By exploiting the DMG, we plan solutions that move the grasp from an initial configuration to the desired one. This solution can include both in-hand manipulation and regrasping. Finally, the sequence of motions required to execute the planned solution is executed by a dual-arm system.

This work is an extension of~\cite{cruciani_dexterous_manipulation_graph}. The improvements are:
\begin{itemize}
	\item A more detailed analysis of the DMG and an in-depth explanation of the DMG search, which was not extensively detailed in our previous work.
	\item The addition of a regrasp action, which enables many more complex manipulation tasks. This regrasp action connects different components of the Dexterous Manipulation Graph, providing solutions in cases in which solely in-hand manipulation would fail, and it is the main improvement with respect to our previous work.
	\item New experiments with a different robot platform, which show the capabilities of the DMG for dexterous manipulation planning by exploiting both in-hand manipulation and regrasping.
\end{itemize}


This paper is organized as follows: in section~\ref{sec_related_work} we summarize the related work in the field of dexterous manipulation, with special focus on extrinsic dexterity; we also analyze the use of dual-arm robots for object manipulation. Section~\ref{sec_dmg_definition} provides an overview of our system and it defines the Dexterous Manipulation Graph and the manipulation primitives that are used for generating it. The process of obtaining the DMG for a certain object's shape is described in section~\ref{sec_graph_generation}. Section~\ref{sec_object_manipulation_planning} shows how to exploit the DMG for planning in-hand manipulation and section~\ref{sec_regrasp_planning} extends this plan to include the possibility of regrasping the object. The motion of the dual-arm robot is described in section~\ref{sec_execution}. Examples of DMG and experiments with the robot are shown in section~\ref{sec_experiments}. Finally, we draw our conclusions and discuss future work in section~\ref{sec_conclusions}.

\section{Related Work}\label{sec_related_work}
This work spans over three important areas: dexterous manipulation, extrinsic dexterity and dual-arm manipulation.

\subsection{Dexterous Manipulation}

In many applications, the robot is required to execute a specific grasp on an object, e.g. if the object is to be used as a tool, or when placing an object in a confined space.
Despite the abundance of work in grasping~\cite{bohg_data_driven_grasp}, achieving a desired grasp on an object is not straightforward: errors in perception and in actuation, as well as limited reachability, lead the execution of a planned grasp to result in a different configuration. To address these, it is possible to place the object back on the supporting surface and apply a new grasp again until the grasp coincides with the desired one~\cite{tournassoud_regrasping}. However, this process requires a sequence of many pick-and-place operations for the purpose of grasping an object, resulting in an inefficient solution. Alternatively, fast regrasping can be executed by tossing the object in the air and adjust the pose of the hand to catch it in the desired configuration~\cite{furukawa_dynamic_regrasping}, but this solution is often infeasible due to the requirements for high-speed object tracking and hand actuation.

Dexterous In-Hand Manipulation can be seen as an alternative to regrasping. Instead of releasing the object and grasping it again, the grasp is adjusted by executing different hand motions that lead to the desired configuration. The ability of the hand to translate or rotate the object is dependent on the dexterity of the hand.
An excellent example of a dexterous hand is the human hand. It remains extremely challenging to build an artificial hand that can mimic its skills~\cite{santina_pisa_soft_hand, or_interpolation_control_posture_in-hand}, both from mechanical design and control perspectives~\cite{bicchi_dexterous_manipulation_review, okamura_dexterous_manipulation_overview,  ozawa_dexterity_control_survey}. 
As an alternative to build complex dexterous hands, some works focus on building custom grippers for specific tasks, lowering the dexterity of the system to the movement needed for the target problem~\cite{bircher_2fingered_gripper_for_reorientation, rojas_underactuated_hand_for_in-hand, rahman_dexterous_gripper, liarokapis_dexterous_adaptive_hands, chavan-dafle_shape-shifting_gripper}.

In addition to the aforementioned challenges, there are also challenges that arise when the end-effectors are used for planning and executing in-hand manipulation. The latter challenges have been addressed in various ways.


The Hierarchical Fingertip Space approach~\cite{hang_HFTS} (HFTS) provides a system for grasping the object and successively adjusting the pose of the fingers to increase the stability of the grasp. This work subdivides the object into small areas to find suitable contacts for the fingers, a concept similar to what we use for generating the Dexterous Manipulation Graph. Differently from our work, the main focus of HFTS is to achieve a stable grasp, and not dexterous in-hand manipulation.

The authors in~\cite{sundaralingam_in-grasp_manipulation} focus on in-grasp manipulation. Their method, called ``relaxed rigidity constraint" is designed to adjust the pose of the object inside the hand. However, it can achieve only small changes in the object's pose, and the contact points are not slid along the surface, but kept stable. Similarly, in~\cite{psomopoulou_stable_pinching} the grasp configuration is changed by moving the fingers' joints and not by changing the contact points on the object. In addition to changing the fingers' joint configurations, the dexterous manipulation planner in~\cite{saut_dexterous_manipulation_planning_roadmaps} proposes a roadmap to plan regrasping as well: one finger can release the contact and move to a new one, while the other fingers keep the object stable. Thus, this planner also places an important assumption, which is that the fingertips cannot slide or rotate on the object's surface.

Recent progress in Reinforcement Learning demonstrated the possibility of applying it to the complex task of object manipulation with a multi-fingered hand. However, despite successfully exploiting a system simulator to learn policies that can be used on a real system, these policies cannot yet be well generalized for different objects~\cite{andrychowicz_learning_dexterous_manipulation}.

\subsection{Extrinsic Dexterity}

In contrast to the works presented in the previous section, several studies address the problem of achieving in-hand manipulation under limited dexterity. This includes our work, which focuses on in-hand manipulation with parallel grippers. With end-effectors that are poor in DOF, the in-hand manipulation task can be achieved by exploiting extrinsic dexterity~\cite{chavan-dafle_extrinsic_dexterity}.

Limited dexterity often results in limited in-hand repositioning capabilities, even when the dexterity is enhanced by means of external supports. Therefore, many works focus on simple in-hand manipulations, that still allow for rearrangements of the object pose inside the gripper. An example of this simple manipulation is \emph{pivoting}. Pivoting is the rotation of the object between the two fingertips, and it admits reorientation along a single axis. 
This in-hand rotation can be initiated by exploiting gravity or inertial forces\cite{sintov_swing-up_regrasping, cruciani_3stages_pivoting, cruciani_integrated_pivoting}. An example of gravity based pivoting is the work in~\cite{vina_adaptive_control_pivoting}, which adopts an adaptive control solution to compensate for the modeling uncertanties of the friction at the pivoting point. The work in~\cite{antonova_RL_pivoting} exploits inertial forces generated by the robot arm for rotating the object, and it uses a Reinforcement Learning approach with noise in the training data to compensate for the friction uncertainty.
Pivoting is an in-hand manipulation strategy that only allows for limited repositioning. However, it can be combined with other strategies, such as \emph{sliding}~\cite{shi_dynamic_sliding, shi_dynamic_sliding_journal}, and provide a more complete range of possibilities. In fact, in our solution, we propose to combine rotational and translational motion, which are treated separately, to achieve the in-hand manipulation goal. 

An additional support that enhances the dexterity, other than gravity, friction and inertial forces, is the use of external contacts with other surfaces~\cite{eppner_exploitation_environmental_consraints_in_grasping, hou_fast_planning_for3D_any-pose-reorientation, almeida_dexterous_manipulation_external_contacts}. The presence of external contacts helps in manipulating the object and allows the robot to reorient it and obtain the desired pose. The work in~\cite{chavan-dafle_sampling-based_planner}, and extended in~\cite{chavan-dafle_in-hand_manipulation_motion_cones}, provides a plan for in-hand manipulation as a sequence of pushes against external fixtures. The plan is based on RRT* and it needs an accurate inverse dynamic solver to predict how the object will slide and rotate within the gripper. 

Our approach provides an efficient graph structure, which integrates object representation and manipulation planning. Instead of relying on external fixtures, we propose to exploit the redundancy of a dual-arm system. Therefore, we do not rely on environment models or on properly placed external fixtures, and the dual-arm system provides additional control over how an object can be pushed inside the gripper. Moreover, it allows the robot to overcome limitations of in-hand manipulation by exploiting regrasping, without the need for planning object placements.

\subsection{Dual Arm Manipulation}

Dual arm systems can be used to manipulate the object using two grippers at the same time. While there are several studies in coordinating dual arm systems and exploiting them for grasping and regrasping~\cite{smith_dualarm_survey}, these systems are not commonly used in the context of in-hand manipulation.

Often, dual-arm manipulation is used for large object handling, and it takes advantage of whole body contacts. For instance, the dual-arm manipulation method proposed in~\cite{murooka_whole-body_manipulation} is formulated using a transition graph that considers possible contacts with the whole body of a humanoid robot. This transition graph is generated by testing the outcome of basic operations, such as lifting or sliding, and a dynamic model of object and environment is required.
The bimanual grasp planner in~\cite{vahrenkamp_bimanual_grasp_planning} is also used for handling large objects and not for fine int-hand manipulation. In~\cite{almeida_cooperative_manipulation} the authors apply a dual-arm system to the manipulation of articulated objects. This application can be seen as an instance of dexterous manipulation, but the contacts between the gripper and the object are always kept fixed, while we focus on adjusting the grasp configuration, which includes changing the position of the contact points.

An example in which a dual arm system is used to grasp an object multiple times is presented in~\cite{sommer_bimanual_tactile_eploration}. However, the goal is to perform tactile exploration to identify the object and later grasp it, and the robot's exploration is passive: a human moves the robot's hand on the object's surface.
The \emph{IK switch} move proposed in~\cite{xian_closed-chain_dualarm_manipulation} is designed to manipulate objects by performing multiple grasps. However, the possibility of sliding or rotating the object while maintaining contact is not taken into account, and the use of regrasping is meant to ease the planning of the two robot manipulators.
Similarly, the work in~\cite{vahrenkamp_dual_arm_manipulation_regrasping}, which focuses on humanoid systems with multi-fingered hands, is used to select between different grasping and regrasping poses that have been predefined, and there is no in-hand manipulation possibility between different grasps. 

The planning proposed in~\cite{saut_two_hand_regrasping} addresses the problem of two-hand regrasping, in which one hand grasps an object from the other hand. We exploit this concept to enhance the manipulability of the object when the sole in-hand manipulation is not enough. However, in this work the authors use multi-fingered hands and they assume that once the object is grasped or handed over its pose inside the hand cannot change. In contrast, we exploit the DMG to achieve more regrasping possibilities for the dual-arm system.

The preparatory manipulation planner described in~\cite{wan_preparatory_manipulation_planning} exploits a \emph{regrasp graph} to plan a solution to regrasp an object in the proper configuration. This graph contains grasp configurations and object poses, and it is different from our Dexterous Manipulation Graph, which instead carries the information about fingers moving along the object's surface. This regrasp graph is used to plan a sequence of pick-and-place and handovers for regrasping to allow the robot to grasp the object as desired. However, the sequence of regrasps on the object does not take into account the possibility of in-hand manipulation, and all the grasp poses must be pre-computed off-line to build the graph.

While it is not always true that using two arms for regrasping is better than a single arm placing the object and picking it up again, dual-arm regrasping is more flexible and it provides a higher success rate; however, single arm regrasp performs better when the grasp and the regrasp pose overlap~\cite{wan_comparing_singlearm_dualarm_regrasp}. In our case, we overcome this complication by exploiting in-hand manipulation. 

To achieve in-hand manipulation with dual arm, we assume that the robot is holding the object with one gripper and the other one is used as additional support to enhance the dexterity. We use the Extended Cooperative Task Space (ECTS)~\cite{park_ECTS} to control the motion of the two arms. The ECTS task specification~\cite{park_ECTS_performance_eval} explores different manipulation scenarios, but it does not address the problem of in-hand manipulation. We obtain the desired relative motion between the grippers given the desired in-hand motion of the object, and the robot's arms are moved accordingly.

\section{Definitions}\label{sec_dmg_definition}
In this section, we define the Dexterous Manipulation Graph and the manipulation types that we use to plan and execute grasp reconfigurations

\subsection{DMG definition}

The Dexterous Manipulation Graph (DMG) is a structure proper to each object, and it depends on the object's shape and partially on the length of the considered gripper's finger. It is a disconnected undirected graph that represents the possible motions of a finger along the object's surface. This graph is defined by a set of nodes $N$ and a set of edges $E$ between these nodes.

\subsubsection{Graph Nodes}

A node $n_{ij}\in N$ represents a configuration of the finger on the object's surface, containing information on both position and orientation. More specifically, each node is a tuple $n_{ij}{=}\langle\textbf{p}_i, A_{ij}\rangle$; $\textbf{p}_i$ is the position of the contact point between the fingertip and the object, and $A_{ij}$, named angular component, is a continuous set of orientations along which the finger in contact at $\textbf{p}_i$ can rotate without collisions. As shown in the example in Fig.~\ref{fig_angle_components}, more nodes in the graph can correspond to the same contact point, depending on the shape of the object. In the example, the point $\textbf{p}_i$ corresponds to two nodes $n_{ij}$ and $n_{ih}$, because when the finger is in contact at this point it can rotate continuously in the set of angles $A_{ij}$ or in the set of angles $A_{ih}$.

\subsubsection{Graph Edges}
\begin{figure}
	\centering
	\includegraphics[width=0.26\textwidth]{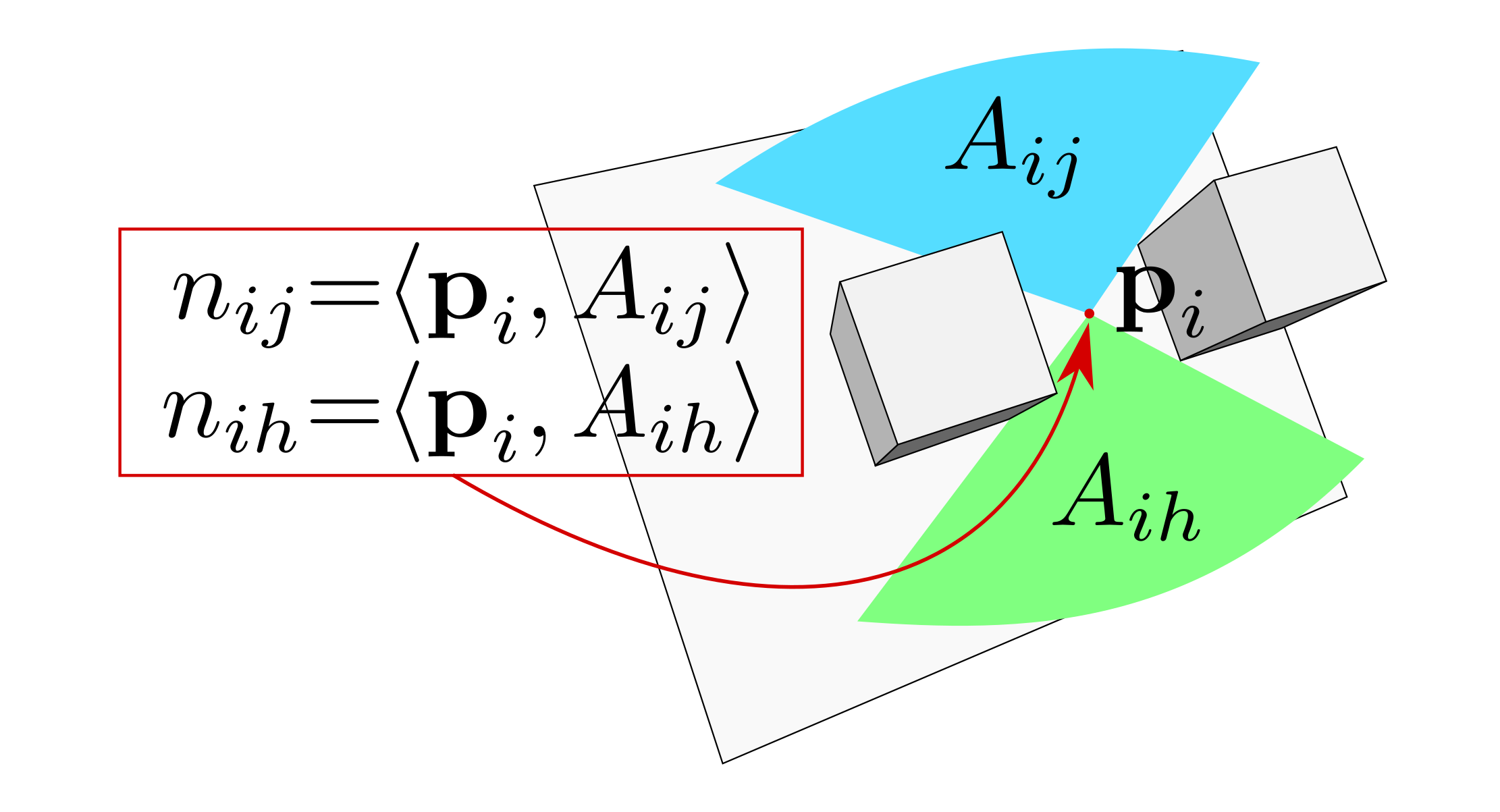}
	\caption{An example of two nodes in the DMG, $n_{ij}$ and $n_{ih}$, that correspond to the same contact point $\textbf{p}_i$, but have two different angular components, $A_{ij}$ and $A_{ih}$.}\label{fig_angle_components}
\end{figure}

An edge $e_{n_{ij}n_{gh}}\in E$ connects two nodes $n_{ij}$ and $n_{gh}$ if it is possible to move the finger along the object's surface between the two configurations. The possible motions taken into account, called in-hand manipulation primitives, are described in section~\ref{sec_in-hand_manipulation_primitives}. In the example in Fig.~\ref{fig_angle_components}, the two nodes $n_{ij}$ and $n_{ih}$ are not connected by an edge because it is not possible to directly move the finger between the two corresponding configurations. In fact, a finger in contact at the point $\textbf{p}_i$ and with an angle $\phi_j\in A_{ij}$ cannot move to an angle $\phi_h \in A_{ih}$ while maintaining the same fingertip contact point, due to collisions with the object.

\subsubsection{Graph Components}

The DMG is a disconnected graph, i.e. it has several connected components that are disconnected from each other. This is due to the nature of the graph, which describes the motions of a gripper's finger on the object's surface: it is not possible for the finger to freely move from any point of the object's surface to another; for instance, a finger cannot move across sharp edges without releasing the contact.
Each connected component $C_i\subseteq N$ contains all the nodes between which a finger can move using the in-hand manipulation primitives. Therefore, it is not possible to move the finger between two nodes $n_{ij}$ and $n_{gh}$ if $n_{ij}\in C_i$ and $n_{gh}\in C_g$, with $C_i\neq C_g$, without releasing the contact with the object's surface. We exploit the different connected components for planning regrasps when an in-hand motion is not possible.

To build the DMG, the object's shape is analyzed according to how the gripper's fingers can move along its surface. In the following, we introduce the manipulation primitives that are considered to describe the fingers' motion. Furthermore, we take into account the possibility of releasing the contact and regrasping in case the desired configuration is not achievable through solely in-hand motions.

\subsection{In-Hand Manipulation Primitives} \label{sec_in-hand_manipulation_primitives}

The in-hand manipulation primitives are motions that describe how a single gripper's finger moves with respect to the object. These motions lead the finger on the object towards a different configuration. 

The configuration of the finger on the object is a tuple $c{=}\langle \textbf{p}, \phi \rangle$, in which $\textbf{p}$ is the position of the contact point between the fingertip and the object's surface, and $\phi$ is the current angle of the finger with respect to the object. It is important to notice that a finger configuration is different from a graph node; in fact, a graph node describes a set of configurations, because it includes the set of possible angles that can be assumed at a certain contact point.

For the DMG generation given the object's shape, we consider the following motions:
\subsubsection{Translation} 
(Fig.~\ref{fig_translation}). The contact point between the fingertip and the object slides along the object's surface, while the orientation of the finger with respect to the object does not change. More specifically, a translation moves the finger from a configuration $c{=}\langle \textbf{p}, \phi \rangle$ to a new configuration $c'{=}\langle \textbf{p}', \phi \rangle$.
\subsubsection{Rotation} 
\begin{figure}[b]
	\centering
	\begin{subfigure}[t]{0.33\columnwidth}
		\centering
		\includegraphics[width=0.8\textwidth]{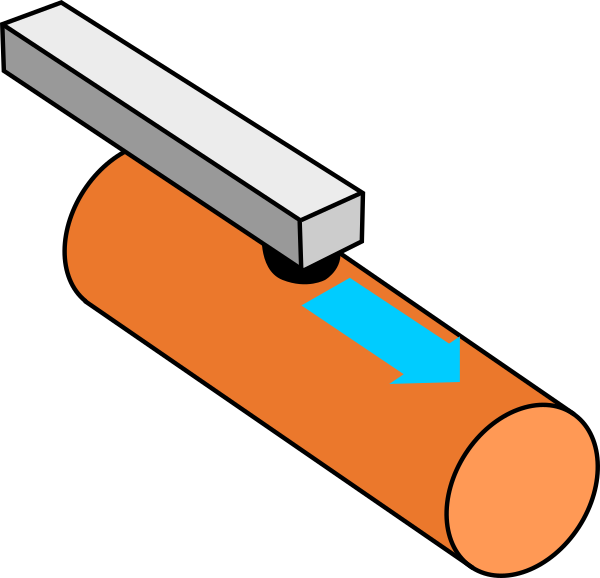}
		\caption{Translation}\label{fig_translation}
	\end{subfigure}%
	\begin{subfigure}[t]{0.33\columnwidth}
		\centering
		\includegraphics[width=0.8\textwidth]{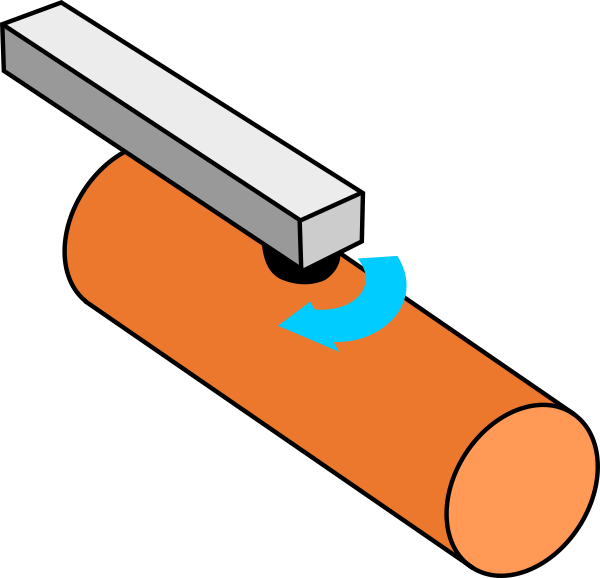}
		\caption{Rotation}\label{fig_rotation}
	\end{subfigure}%
	\begin{subfigure}[t]{0.33\columnwidth}
		\centering
		\includegraphics[width=0.8\textwidth]{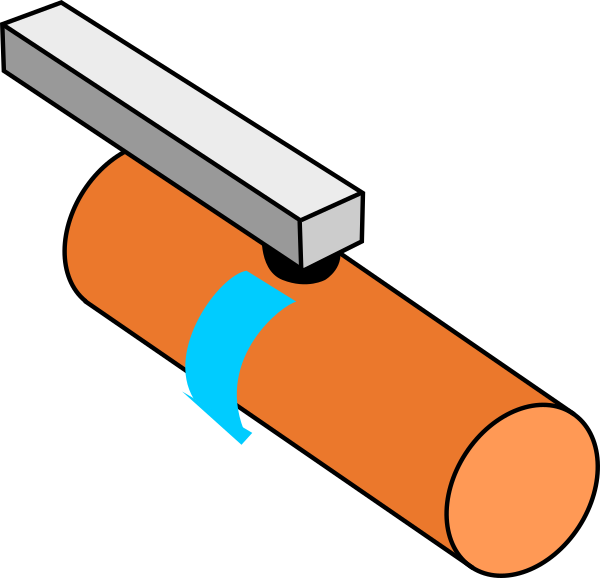}
		\caption{Rototranslation}\label{fig_rototranslation}
	\end{subfigure}
	\caption{The three manipulation primitives that describe the motion of the contact point between the fingertip and the object. The finger of a parallel gripper is represented in gray, and its fingertip in black.}
\end{figure}

(Fig.~\ref{fig_rotation}). The contact point between the fingertip and the object is kept fixed, while the finger's orientation with respect to the object changes. More specifically, a rotation moves the finger from a configuration $c{=}\langle \textbf{p}, \phi \rangle$ to a new configuration $c'{=}\langle \textbf{p}, \phi' \rangle$.

A combination of these two motions at the same time, a \emph{rototranslation} (Fig.~\ref{fig_rototranslation}), is not considered in our current DMG generation because in several cases it requires prehensile grasps to be executed. Therefore, motions that require a simultaneous change in the finger's orientation and in the position of the fingertip's contact point are not considered feasible in our system. Instead, we use non-prehensile pushing to execute either a translation or a rotation.

\subsection{Regrasping}\label{sec_regrasping_primitive}

\begin{figure}
	\centering
	\begin{subfigure}[t]{\columnwidth}
		\centering
		\includegraphics[width=0.7\textwidth]{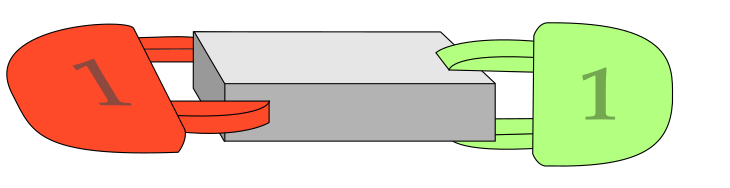}
		\caption{The initial grasp of the gripper on the object is shown in red, on the left, and the desired one in green, on the right. It is not possible to reach the desired grasp by sliding or rotating the fingers on the object's surface because the fingertip contacts lie on areas between which it is not possible to slide.}\label{fig_regrasp1}
	\end{subfigure}
	\begin{subfigure}[t]{\columnwidth}
		\centering
		\includegraphics[width=0.7\textwidth]{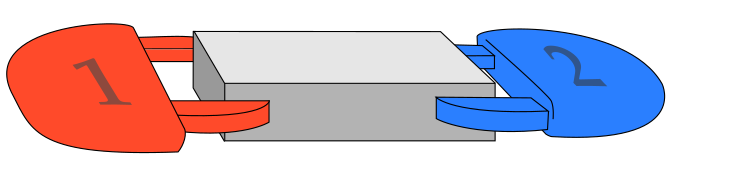}
		\caption{Regrasp is needed. A second gripper, shown in blue, helps holding the object while the first gripper releases it. However, this second gripper prevents the first gripper from regrasping at the desired configuration.}\label{fig_regrasp2}
	\end{subfigure}
	\begin{subfigure}[t]{\columnwidth}
		\centering
		\includegraphics[width=0.7\textwidth]{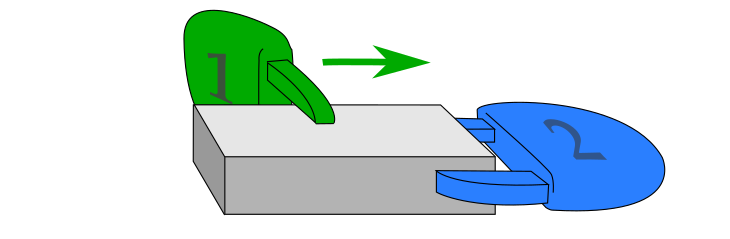}
		\caption{The first gripper can execute a new grasp, shown in dark green, that is not in collision with the second gripper. Once the second gripper releases the object, the first gripper can achieve the desired grasp thanks to in-hand manipulation.}\label{fig_regrasp3}
	\end{subfigure}%
	\caption{A simple example of a desired grasp on an object that requires regrasp and in-hand manipulation.}\label{fig_regrasp}
\end{figure}%

It is possible that a desired grasp is not reachable from a certain initial configuration by using the in-hand primitives, due to the object's shape. In this case, we introduce the possibility of regrasping the object at a different configuration. This new grasp can be the desired grasp, if it is possible to directly regrasp there, or it can lead the gripper closer to the desired grasp to a pose from which subsequent in-hand manipulation movements will produce the desired grasp.

The use of the DMG to plan regrasping helps in manipulating objects of complex shapes for which a direct grasp in the desired configuration is not achievable or difficult to execute. Moreover, since we consider a dual arm robot, the regrasping action takes advantage of the second gripper as an additional support to change the object's configuration, but the second gripper's grasp should also be planned according to the desired grasp. The sequence in Fig.~\ref{fig_regrasp} shows an example in which regrasping is needed to reach the desired configuration.
In a general setting, the grasp and release actions can also need in-hand manipulation depending on the shape of the object, especially if it presents concavities. Therefore, we exploit the combination of regrasping with the DMG representation to achieve flexible grasp reconfigurations.

The regrasping action allows the gripper's finger to move into the component of the DMG that contains the desired grasp. That is, the regrasping action has the role of creating bridges between the disconnected sub-graphs that compose the DMG.



\section{Graph Generation}\label{sec_graph_generation}
This section describes the process of obtaining the DMG starting from the object's shape. It corresponds to step 1 in Fig.~\ref{fig_overview}.

\subsection{Overview}

The main idea for obtaining the DMG is to divide the object's surface into small areas, corresponding to contact points with the fingertip, and analyze how the gripper's finger can move between them. The size of the area depends on the chosen approximation of the fingertip contact with respect to the size of the object.

The connection between an area an the neighboring areas is refined based on the possibility for a fingertip to move from one to the other using the in-hand manipulation primitives defined in section~\ref{sec_in-hand_manipulation_primitives}. To each area, one or more nodes in the DMG are associated; these nodes contain information about the fingertip contact, identified by the area's centroid, and the possible range of orientations (angular components) that the finger can assume at that contact point. The detailed process of DMG generation is described in the next session and summarized in Algorithm~\ref{algorithm_DMG_generation}. 


\subsection{From Object's surface to DMG}\label{sec_from_obj_to_dmg}

\begin{algorithm}[t]
	\small
	
	\SetNoFillComment
	\DontPrintSemicolon
	\SetKwInOut{Input}{Input}
	\SetKwInOut{Output}{Output}
	
	\Input{$P$, $l_f$, $\delta_n$, $r_{angle}$}
	\tcc{P is the object represented as point cloud}
	
	$N_s, E_s \leftarrow$supervoxels of P and their adjacency\\
	$N\leftarrow \emptyset$ \tcp*{empty set of nodes}
	$E\leftarrow \emptyset$ \tcp*{empty set of edges}
	$A_{360}\leftarrow\{0:r_{angle}:360\}$ \tcp*{set of discrete angles}
	\ForEach{$e_{a_i a_j}\in E_s$}{
		$\textbf{p}_{ai}\leftarrow$centroid of the area $a_i$\\
		$\textbf{p}_{aj}\leftarrow$centroid of the area $a_j$\\
		$n_{a_i}\leftarrow \langle \textbf{p}_{ai}, A_{360}\rangle$\\
		$n_{a_j}\leftarrow \langle \textbf{p}_{aj}, A_{360}\rangle$\\
		$N\cup\{ n_{a_i}, n_{a_j} \}$\\
		$E \cup \{ e_{n_{a_i} n_{a_j}}\}$
	}
	
	\ForEach{$n_{a}\in N$}{
		$E\leftarrow$translation\_refinement($n_{a}$, $N$, $E$, $\delta_n$)\\	
	}
	
	\ForEach{$n_a\in N$}{
		$A\leftarrow$angular\_component($P$, $l_f$, $n_a$)\\
		\uIf{$A \mathrm{~is~empty}$}{			
			\ForEach{$e_{nm}\in E \mathrm{~s.t.~} n \mathrm{~is~} n_a \mathrm{~or~} m \mathrm{~is~} n_a$}{
				$E\leftarrow E\smallsetminus\{e_{nm}\}$	
			}
			$N\leftarrow N\smallsetminus\{n_a\}$ \tcp*{no contact is possible}
		}
		\Else{
			$N, E\leftarrow$split\_node($n_a$, $A$, $N$, $E$, $r_{angle}$)
		}	
	}
	
	\ForEach{$n_a\in N$}{
		$E\leftarrow$rotation\_refinement($n_a$, $N$, $E$)\\
	}
	
	DMG $\leftarrow\langle N, E \rangle$\\
	
	\Return DMG
	
	\caption{DMG\_generation}\label{algorithm_DMG_generation}
\end{algorithm}

\begin{algorithm}[t]
	\small
	
	\SetNoFillComment
	\DontPrintSemicolon
	\SetKwInOut{Input}{Input}
	\SetKwInOut{Output}{Output}
	\caption{translation\_refinement}\label{algorithm_translation_refinement}
	
	\Input{$n_{a_i}$, $N$, $E$, $\delta_n$}

	\tcc{$a_i$ is the supervoxel corresponding to node $n_{a_i}$}
	$\hat{\textbf{n}}_{a_i}\leftarrow$normal to the surface in $a_i$\\
	\ForEach{ $n_{a_j}\in N$ \upshape{connected to} $n_{a_i}$}{
		$\hat{\textbf{n}}_{a_j}\leftarrow$normal to the surface in $a_j$\\
		\If{\upshape{$\parallel \hat{\textbf{n}}_{a_i} - \hat{\textbf{n}}_{a_j}\parallel_2 > \delta_n$}}{
			$E\leftarrow E \smallsetminus \{e_{n_{a_i} n_{a_j}}\}$
		}
	}
	\Return $E$
\end{algorithm}

\begin{algorithm}[t]
	\small
	
	\SetNoFillComment
	\DontPrintSemicolon
	\SetKwInOut{Input}{Input}
	\SetKwInOut{Output}{Output}
	
	\Input{$P$, $l_f$, $n_a$}
	
	$\textbf{p}_{a}\leftarrow$centroid of $n_a$\\
	$A_{360}\leftarrow$ set of angles of $n_a$\\
	$A\leftarrow \emptyset$\\
	$\phi\leftarrow0$\\
	\ForEach{\upshape{$\phi\in A_{360}$}}{
		\If{\upshape{finger of length $l_f$ in contact at $\textbf{p}_{a}$ with angle $\phi$ not in collision with object $P$}}{
			$A\leftarrow A\cup\{\phi\}$
			
		}
	}
	
	\Return $A$
	
	\caption{angular\_component}\label{algorithm_angle_component}
\end{algorithm}

\begin{algorithm}[t]
	\small
	
	\SetNoFillComment
	\DontPrintSemicolon
	\SetKwInOut{Input}{Input}
	\SetKwInOut{Output}{Output}
	
	\Input{$n_a$, $A$, $N$, $E$, $r_{angle}$}
	$K_{angle}\leftarrow\lceil \frac{360}{r_{angle}}\rceil$\\
	$\phi_{max}\leftarrow K_{angle}\;r_{angle}$\\
	$k\leftarrow 0$\\
	$A_a\leftarrow\emptyset$ \tcp*{empty set of angles}
	$A_A\leftarrow\emptyset$ \tcp*{empty set of sets}
	\While{$k<K_{angle}{-}1$}{
		\tcp{the set $A$ is ordered}
		$\phi_k\leftarrow k$-th element of $A$\\
		$A_a\leftarrow A_a\cup\{ \phi_k \}$\\
		$\phi_{k{+}1}\leftarrow k{+}1$-th element of $A$\\
		\If{$\lvert \phi_{k{+}1}-\phi_k \rvert>r_{angle}$}{
			$A_A\cup \{ A_a\}$\\
			$A_a\leftarrow\emptyset$
		}
		
		$k\leftarrow k$+$1$
	}
	\uIf{$\phi_{max}\in A_a$}{
		$A_0\leftarrow$first set in $A_A$\\
		\If{$0\in A_0$}{
			$A_A\leftarrow A_A \smallsetminus \{ A_0\}$\\
			$A_0\leftarrow A_0 \cup A_a$\\
			$A_A\leftarrow A_A \cup \{A_0\}$\\
		}
	}
	\Else{
		$A_A\cup \{ A_a\}$
	}
	$\textbf{p}_a\leftarrow$ centroid of $n_a$\\
	\ForEach{$A_i\in A_A$}{
		$n_{ai}\leftarrow \langle \textbf{p}_a, A_i \rangle$\\
		$N\leftarrow N \cup \{n_{ai}\}$\\
		\ForEach{\upshape{$e_{nm}\in E$ s.t. $n$ is $n_a$}}{
			$E\leftarrow E\cup\{ e_{n_{ai}m}\}$\\
		}
		\ForEach{\upshape{$e_{nm}\in E$ s.t. $m$ is $n_a$}}{
			$E\leftarrow E\cup\{ e_{n n_{ai}}\}$\\
		}
	}
		
	\ForEach{\upshape{$e_{nm}\in E$ s.t. $n$ is $n_a$ or $m$ is $n_a$}}{
		$E\leftarrow E\smallsetminus\{e_{nm}\}$	
	}
	$N\leftarrow N\smallsetminus\{n_a\}$\\

	\Return $N$, $E$
	
	\caption{split\_node}\label{algorithm_split_node}
\end{algorithm}

\begin{algorithm}[t]
	\small
	
	\SetNoFillComment
	\DontPrintSemicolon
	\SetKwInOut{Input}{Input}
	\SetKwInOut{Output}{Output}
	
	\Input{$n_{a_i}$, $N$, $E$}
	
	$A_{a_i}\leftarrow$ angular component of $n_{a_i}$\\
	\ForEach{\upshape{$n_{a_j}\in N$ connected to $n_{a_i}$}}{
		$A_{a_j}\leftarrow$ angular component of $n_{a_j}$\\
		\If{\upshape{$A_{a_i}\cap A_{a_j}$ is empty}}{
			$E\leftarrow E\smallsetminus\{e_{n_{a_i}n_{a_j}}\}$\\
		}
	}
	
	\Return $E$
	
	\caption{rotation\_refinement}\label{algorithm_rotation_refinement}
\end{algorithm}

We represent the object as a point cloud and we subdivide its surface into smaller areas of approximate resolution $r_{area}$ using the method of supervoxels~\cite{papon_voxel_cloud_connectivity}. Each area is segmented out from the rest of the object given the desired resolution and according to common characteristics, such as geometric information.

An area $a$, called supervoxel, is associated to its centroid $\textbf{p}_a$ and its normal $\hat{\textbf{n}}_a$. The geometric proximity of an area with its neighbor areas is contained in a graph. This graph is described by a set of nodes $N_s$ and a set of edges $E_s$. Each node corresponds to a supervoxel and an edge connects a supervoxel to its neighbors. 

The connected graph describing the object's surface subdivided into areas is the starting point for the generation of the disconnected DMG. An initial set of nodes $N$ and an initial set of edges $E$ replicates the structure of the supervoxels graph. However, the new set of nodes differs from $N_s$ because a node is not representing an area, but it has the same components of a DMG node.

A node $n_a\in N$, corresponding to the supervoxel $a$, is a tuple $n{=}\langle \textbf{p}_a, A_{360}\rangle$. $A_{360}$ is a discrete set that contains all the possible orientations of the finger in the range between $0^\circ$ and $360^\circ$. We choose to represent the possible orientation with resolution $r_{angle}\in \mathrm{I\!R}$, so that the set $A_{360}$ contains all the $K_{angle}$ multiples of $r_{angle}$ within $[0, 360)$. 
Assuming the elements of the set organized in increasing order, the first element is $\phi_0{=}0$ and the last one is $\phi_{K_{angle}{-}1}{=}K_{angle}\:r_{angle}$.
Therefore, at this step each node in the graph corresponds to a fingertip contact point that is the centroid of a supervoxel, with the assumption that a finger in contact can freely rotate between all angles.
The initial connected graph described by $N$ and $E$ is then refined to obtain the DMG.

First, we execute a \emph{translation refinement}, summarized in Algorithm~\ref{algorithm_translation_refinement}. This refinement removes the connections between areas if a fingertip cannot translate between them. 

Two nodes $n_{a_i}$ and $n_{a_j}$, connected by the edge $e_{n_{a_i} n_{a_j}}$, correspond to the two areas $a_i$ and $a_j$. A fingertip cannot translate between these two areas if
\begin{equation} \label{eq_normal_threshold}
	\parallel \hat{\textbf{n}}_{a_i} - \hat{\textbf{n}}_{a_j}\parallel_2 > \delta_n,
\end{equation}
where $\delta_n$ is a threshold used to ensure that a fingertip will not move across sharp edges. This threshold is adjusted to allow motions along non-flat surfaces to the desired degree. 

If the condition in (\ref{eq_normal_threshold}) is verified, the edge $e_{n_{a_i} n_{a_j}}$ is removed from the graph. This removal can break the graph into disconnected components, which correspond to different faces of the object. In case of curved surfaces, the disconnection between the areas is subject to the chosen tolerance and it should be chosen to ensure the feasibility of the translation given the available system.

To each disconnected component, we associate a reference frame. The average normal $\hat{\textbf{n}}_c$ between all the normals of the nodes belonging to this component corresponds to the x axis of this reference frame; the y axis is taken as an orthogonal vector to the average normal and the z axis is the result of the cross product between the previous two axes. This reference frame is used to describe the finger's orientation as an angle in the yz plane with a common reference between all the connected nodes of the component.

The following step, summarized in Algorithm~\ref{algorithm_angle_component}, consists in identifying the admissible rotations for the finger. 
For each node $n_a$, the finger in contact at the point $\textbf{p}_a$ is checked for possible collisions with the object. More specifically, the finger of given length $l_f$ is rotated at each angle $\phi\in A_{360}$ and the angle is removed from the set if it is in collision. The angle set associated with $n_a$ now contains only the collision-free angles that the finger can assume when its fingertip is in contact with the object in $\textbf{p}_a$. If this set is empty, then this contact cannot be achieved and the node $n_a$ is removed from the graph along with all its edges.

Given an angle $\phi_k$ and the subsequent angle in the set $\phi_{k+1}$, the finger cannot rotate between the two if
\begin{equation}\label{eq_continuous_angles}
	\lvert \phi_{k+1}-\phi_k \rvert>r_{angle}.
\end{equation}
 The only exception is between the angles $\phi_0$ and $\phi_{K_{angle}{-}1}$, because it is possible to rotate between the two even though (\ref{eq_continuous_angles}) is verified.

If there is at least one rotation gap, that is there are two angles in the set between which the finger cannot rotate, this angle set is split into two of more sets $A_{a1}$, $A_{a2}$, ..., $A_{aK_{set}}$. Each of these sets contains a subset of the admissible angles for a finger in contact at $\textbf{p}_a$, between which it is possible to rotate without incurring in collisions. 

When $K_{set}{>}1$, the area $a$ of the object must correspond to more than one node in the DMG. Therefore, the node $n_a$ is removed from the graph and it is split into different nodes $n_{a1}{=}\langle \textbf{p}_a, A_{a1} \rangle$, $ n_{a2}{=}\langle \textbf{p}_a, A_{a2} \rangle$, ..., $n_{aK_{set}}{=}\langle \textbf{p}_a, A_{aK_{set}} \rangle$, each one containing a different set of sequential angles. To each of these nodes we introduce new edges that connect it to the same nodes to which $n_a$ was connected. However, there is no edge that connects these new nodes to each other. This process of splitting one node into multiple nodes according to the angular components is summarized in Algorithm~\ref{algorithm_split_node}.

Finally, the graph with the new nodes refined according to the \emph{rotation refinement} summarized in Algorithm~\ref{algorithm_rotation_refinement}. For each edge $e_{n_{ij} n_{gh}}$, the two angle sets $A_{ij}$ and $A_{gh}$ of the two connected nodes are analyzed. In fact, if $A_{ij}\cap A_{gh}{=} \emptyset$, it is not possible for the finger in contact at $\textbf{p}_i$ to move to the contact point of the other node $\textbf{p}_g$ and vice-versa, because there is not a common admissible orientation. Therefore, the edge $e_{n_{ij} n_{gh}}$ is removed from the graph. 


After the translation and the rotation refinements, the obtained graph is the DMG. This graph describes how a finger can move from one node to the other according to the motions described in section~\ref{sec_in-hand_manipulation_primitives}. 

The obtained graph is disconnected.
We can identify the subsets of nodes that belong to the same component, for instance using breadth-first search on all the nodes in $N$ and identifying a new component whenever the search ends without having visited all the nodes. Therefore, we can associate to the DMG a set of components $C{=}\{C_1, C_2, ..., C_{K_{comp}}\}$. These components describe the subset of nodes between which it is possible to move; in fact, it is not possible to move between nodes that belong to different components, as there is no edge that connects them.

\section{In-Hand Manipulation Planning}
\label{sec_object_manipulation_planning}
This section explains how to use the DMG to plan in-hand manipulation and move the object from an initial grasp to the desired one without releasing contact between fingertips and object. This procedure is the first part of step 2 in Fig.~\ref{fig_overview}.

\subsection{Grasp Configuration Description}

The goal of object manipulation planning is to find the sequence of manipulation primitives that leads the object from an initial grasp configuration to the desired one. We consider in-hand manipulation with a parallel gripper. The grasp of the gripper on the object can be described by the contact point of the two fingertips on the object $\textbf{p}_{f_1}$ and $\textbf{p}_{f_2}$ and by the orientation of the fingers $\textbf{q}_f$, described as a quaternion, which is the same for both due to symmetry.

The points $\textbf{p}_{f_1}$ and $\textbf{p}_{f_2}$ lie on the object's surface; therefore, they belong to one of the supervoxels used to compute the DMG and they can easily be associated with the corresponding nodes that contain the centroid of the supervoxel. By using the reference frame associated to the connected component in the DMG, mentioned in section~\ref{sec_from_obj_to_dmg}, the orientation of the finger in space is translated into an angle in the yz plane of this frame. Using this angle, the fingers are associated to the start configurations $c_{s_1}{=}\langle \textbf{p}_{s_1}, \phi_{s_1} \rangle$ and $c_{s_2}{=}\langle \textbf{p}_{s_2}, \phi_{s_2} \rangle$, where $\textbf{p}_{s_1}$ and $\textbf{p}_{s_1}$ are the centroids of the supervoxels containing the fingertips' contact points. Similarly, the desired contact points for the first and second finger $\textbf{p}_{f_1}'$ and $\textbf{p}_{f_2}'$ and the desired orientation $\textbf{q}_f'$ can be described using the configurations $c_{d_1}{=}\langle \textbf{p}_{d_1}, \phi_{d_1} \rangle$ and $c_{d_2}{=}\langle \textbf{p}_{d_2}, \phi_{d_2} \rangle$.

\subsection{Search for In-Hand Path}\label{sec_search_for_in-hand_path}

Given the initial grasp, described by the fingers' configurations $c_{s_1}$ and $c_{s_2}$, and the desired grasp, $c_{d_1}$ and $c_{d_2}$, the in-hand manipulation planning finds a sequence of in-hand manipulation primitives that moves the object inside the gripper from the initial to the desired configuration.

Each finger configuration is associated to a node in the graph, leading to two initial nodes, $n_{s_1}{=}\langle\textbf{p}_{s_1}, A_{s_1}\rangle$ and $n_{s_2}{=}\langle\textbf{p}_{s_2}, A_{s_2}\rangle$, and to two desired nodes, $n_{d_1}{=}\langle\textbf{p}_{d_1}, A_{d_1}\rangle$ and $n_{d_2}{=}\langle\textbf{p}_{d_2}, A_{d_2}\rangle$. The angular components of these nodes are so that $\phi_{s_1}^r\in A_{s_1}$, $\phi_{s_2}^r\in A_{s_2}$, $\phi_{d_1}^r\in A_{d_1}$ and $\phi_{d_2}^r\in A_{d_2}$. By $\phi^r$ we denote the closest angle to $\phi$ that belongs to the discrete set of angles $A_{360}$ with the chosen resolution $r_{angle}$.

The in-hand manipulation planning is a search in the DMG for a path that connects $n_{s_1}$ with $n_{d_1}$ and $n_{s_1}$ with $n_{d_2}$.

With $C_{s_1}$, $C_{s_2}$, $C_{d_1}$ and $C_{d_2}$ denoting the graph's components containing respectively the two initial nodes and the two desired nodes, there is no in-hand path that can lead from the initial to the desired configuration if
\begin{equation}\label{eq_component_intersection}
	C_{s_1}\neq C_{d_1} \;\;\mathrm{or}\;\; C_{s_2}\neq C_{d_2}.
\end{equation}
In this case, the desired grasp can only be achieved by releasing the contact. The process of releasing and regrasping the object is described in section~\ref{sec_regrasp_planning}. 
If the condition in (\ref{eq_component_intersection}) is not verified, an in-hand path is searched for in the DMG.

First of all, we select one finger as principal finger and the other finger as secondary finger. In fact, the DMG is built considering the motion of only one finger on the object, while we consider the problem of in-hand manipulation with a parallel gripper. Therefore, the search in the DMG must be carried out by following the motions on one finger.

\begin{figure}
	\centering
	\begin{subfigure}[t]{0.5\columnwidth}
		\centering
		\includegraphics[width=0.7\columnwidth]{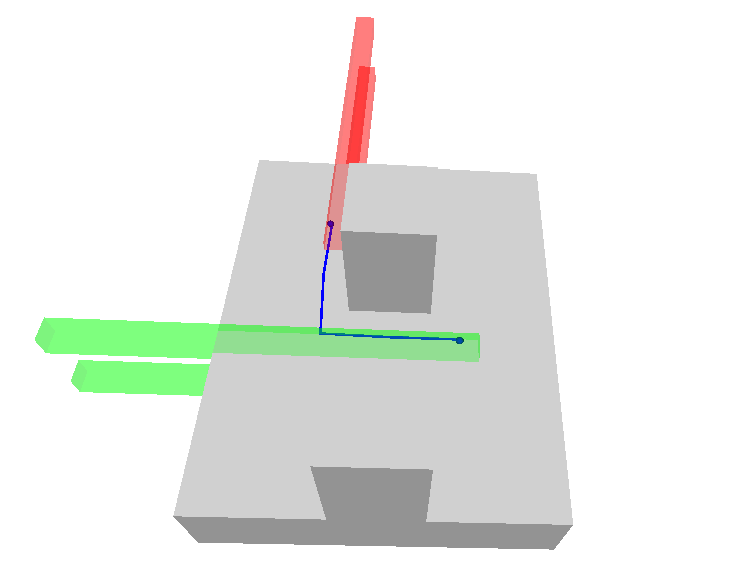}
	\end{subfigure}%
	\begin{subfigure}[t]{0.5\columnwidth}
		\centering
		\includegraphics[width=0.7\columnwidth]{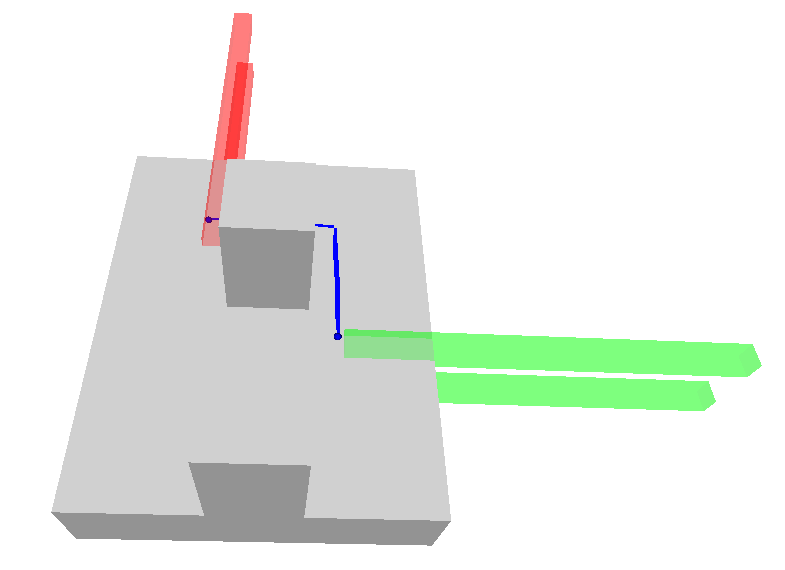}
	\end{subfigure}
	\caption{An example of graph search from the same initial configuration (in red) to the same final contact point, but the desired orientation is different (in green) between the left and the right figure. The DGM's nodes contain angular components, so that the path along the object's surface can be followed by the finger at the given orientation. The object is the same of Fig.~\ref{fig_angle_components}.}\label{fig_angle_component_path}
\end{figure}

Without loss of generality, we assume that $1$ denotes the principal finger and $2$ denotes the secondary finger. The graph search is carried out in the component of the principal finger $C_1$ (equivalent to $C_{s_1}$ and $C_{d_1}$) using Dijkstra's algorithm~\cite{dijkstra}. The inclusion of angular components in the DMG's nodes can result in different path's on the object's surface, even though the initial and final contact points are the same, when the desired orientation differs, as shown in the examples in Fig.~\ref{fig_angle_component_path}.

\begin{figure}
	\centering
	\begin{subfigure}[t]{\columnwidth}
		\centering
		\includegraphics[width=0.85\textwidth]{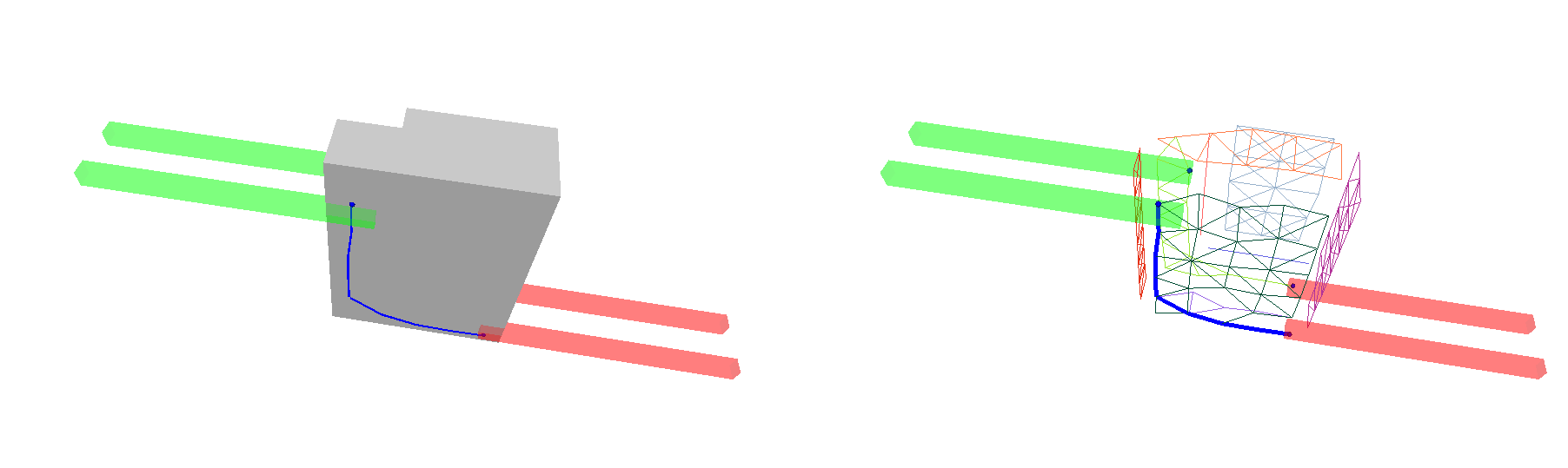}
	\end{subfigure}
	\begin{subfigure}[t]{\columnwidth}
		\centering
		\includegraphics[width=0.8\textwidth]{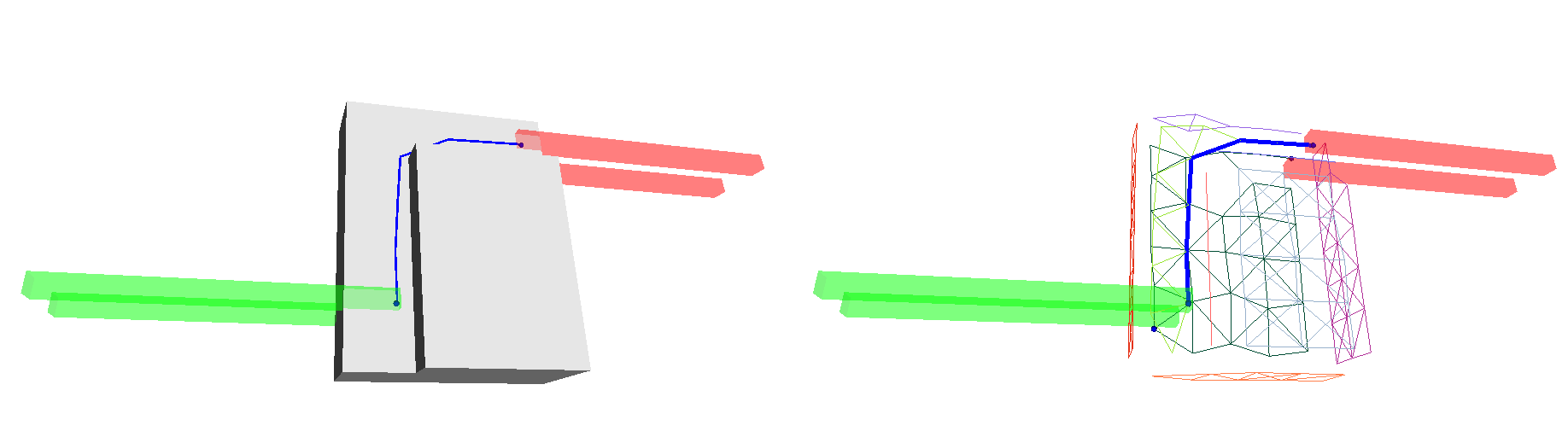}
	\end{subfigure}
	\caption{An example of graph search from the initial grasp (red fingers) to the desired grasp (green fingers). The top and bottom images show the outcome for the same object with a different finger selected as principal finger. In both cases the solution found, shown as blue path, is valid, but it leads to two slightly different in-hand motions. The DMG is shown on the right, and each color corresponds to a different component.}\label{fig_principal_finger}
\end{figure}

During the search, the secondary finger follows the principal finger in the component $C_2$ (equivalent to $C_{s_2}$ and $C_{d_2}$) and it is used to ensure that the followed path is valid for both of the gripper's fingers. As shown in Fig.~\ref{fig_principal_finger}, the choice of one finger rather than the other as principal finger does not affect the validity of the in-hand path found, and hence it can be assigned arbitrarily.

At each iteration of the DMG search, a transition between two connected principal nodes is checked, e.g. from $n_{j_1}$ to $n_{i_1}$.
To a principal node $n_{i_1}\in C_1$ corresponds a secondary node $n_{i_2}$ that represents the secondary finger. This node is found as the node corresponding to the supervoxel that contains the intersection point between the object and the ray from $\textbf{p}_{i_1}$ in the direction opposite to the normal of the area $i_1$. Depending on the object's shape, different situations can occur:
\begin{enumerate}
	\item there is only one intersection point, corresponding to the node $n_{i_2}$. \label{item_one_intersection_one_node}
	\item there is only one intersection point, but it corresponds to multiple nodes $n_{i_2 0}, n_{i_2 1}, ...$ with different angular components. These nodes could also belong to different graph components. \label{item_one_intersection_multiple_nodes}
	\item there are multiple intersection points, corresponding to different nodes $n_{i_2 0}, n_{i_2 1}, ...$ that belong to different graph components $C_{i_2 0}, C_{i_2 1}, ...$\label{item_multiple_intersections_multiple_nodes}
\end{enumerate} 
In the condition of point \ref{item_one_intersection_one_node}, the node $n_{i_2}$ is a valid node for the secondary finger only if the finger can assume the same orientation of the principal finger in this node. The set of possible angles $A_{i_2}$ can be transformed into a new set $A_{i_2}'$ that correspond to the angles in $A_{i_2}$ in the reference frame of $C_1$. Therefore, the secondary node is valid if
\begin{equation}\label{eq_node_validity_angular_condition}
	A_{i_1}\cap A_{i_2}'\neq\emptyset.
\end{equation}
During the search, the node $n_{i_1}$ will be replaced by the node $n_{i_1}'{=}\langle \textbf{p}_{i_1},  A_{i_1}{\cap} A_{i_2}'\rangle$.
In the conditions of points \ref{item_one_intersection_multiple_nodes} and \ref{item_multiple_intersections_multiple_nodes} a node is only valid if it belongs to $C_2$ and it satisfies the condition in (\ref{eq_node_validity_angular_condition}).

\begin{figure}[b]
	\centering
	\includegraphics[width=0.6\columnwidth]{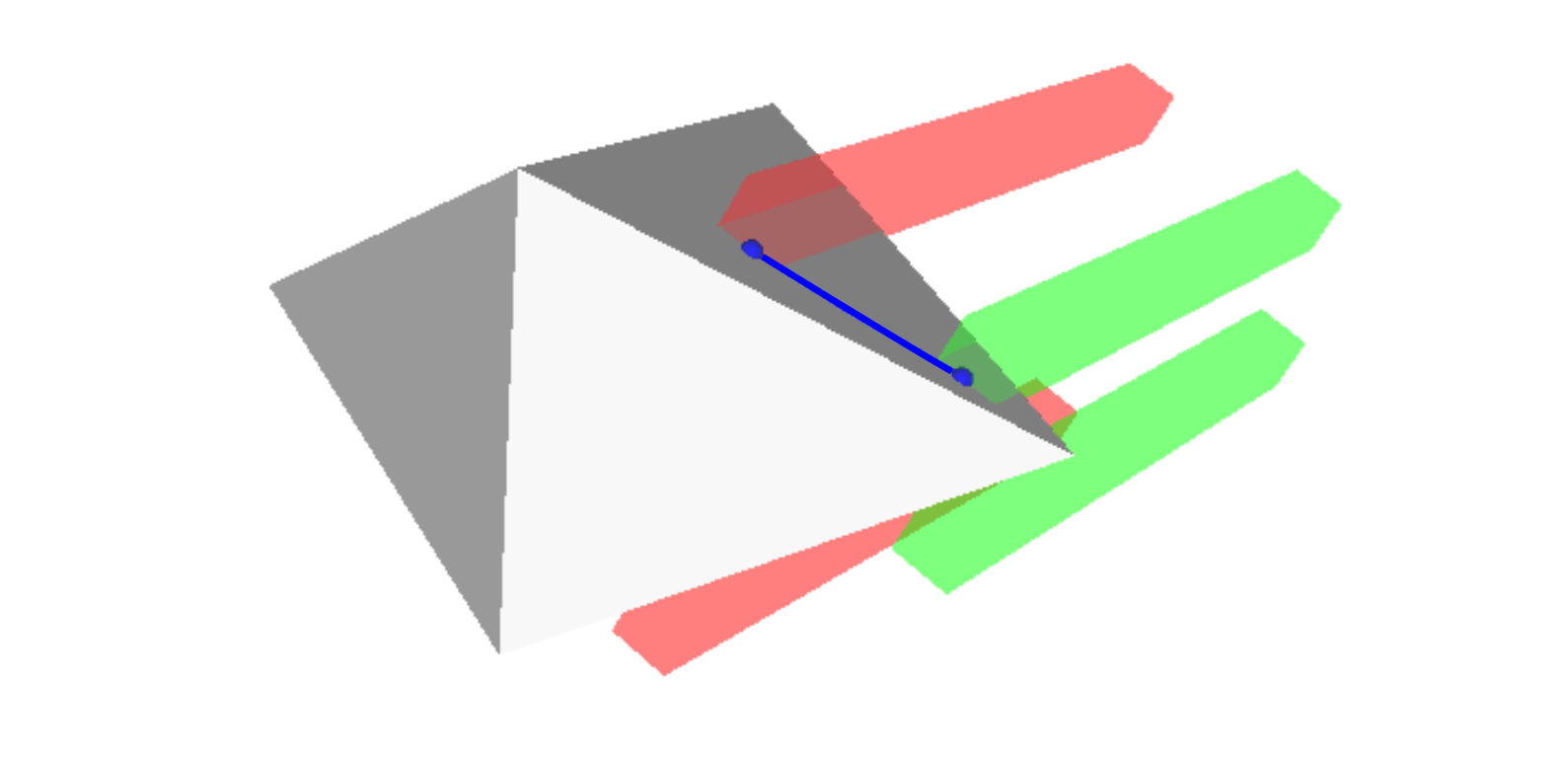}
	\caption{The fingers move from the initial grasp (red) to the desired one (green) by sliding on a prismatic object. This translation requires an adjustment of the fingers' opening because the surfaces are not parallel. This kind of situation can be adjusted depending on the available system and preferences.}\label{fig_non_parallel_surfaces_translation}
\end{figure}

In addition, the difference between the normals in $n_{i_1}$ and $n_{i_2}$ can be checked to allow or not motions along curved or skewed surfaces, such as the example in Fig.~\ref{fig_non_parallel_surfaces_translation}.

If there are no valid nodes, a transition of the principal finger that moves from node $n_{j_1}$ to node $n_{i_1}$ is not allowed. Therefore, the cost of edge $e_{n_{j_1} n_{i_1}}$ is set to ${+}\infty$ to ensure that this path is not chosen.
As a consequence, the path solution may not be the shortest path for the principal finger in its corresponding component, because it must consider the motion of the secondary finger as well.
The right image in Fig.~\ref{fig_principal_finger} shows an example: the finger does not proceed straight to the goal because the opposite finger cannot follow it along that path.

If the node is valid, the cost $\xi_{n_{j_1} n_{i_1}}$ of an edge connecting two nodes $n_{j_1}$ and $n_{i_1}$ can be chosen in different ways.
A first possible choice is
\begin{equation}
	\xi_{n_{j_1} n_{i_1}}=\lVert \textbf{p}_{j_1}-\textbf{p}_{t_1} \rVert_2,
\end{equation}
that is the simple Euclidean distance between the two centroids of the supervoxels associated with the nodes. This distance minimizes the distance in terms of translations of the fingertips on the object's surface.

As an alternative, the cost can be adapted and chosen according to how expensive a certain movement is considered to be. For instance, some translation directions could be more difficult to execute, either due to the object's shape or, as we discuss later in section~\ref{sec_motion_execution_translation}, due to the robot's configuration. Therefore, these directions can be penalized by increasing the cost. 

In addition, if moving from one node to another involves a rotation, the requirement of this rotation in addition to the translation can be considered as more expensive and the cost of the edge increased.
While the association between translation and edge in the graph is straightforward, the rotation is not explicitly represented in the DMG. In fact, a node contains the Cartesian position of the centroid associated with the fingertip contact point, but it contains more than one angle to associate with the finger's orientation.

A solution path in the DMG is a sequence of nodes, and not a sequence of finger configurations. For reasons of notation simplifications, we explain how to obtain the sequence of rotations from a sequence of nodes in section~\ref{sec_sequence_of_angles}. However, this procedure can be executed at the same time of the graph search to allow for the cost of the rotation to be taken into account together with the cost of the translation. For instance, if the finger is with an angle $\phi_{j_1}$ while in the node $n_{j_1}$, and the considered successive node is $n_{i_1}$, a rotation is needed if $\phi_{j_1}\notin A_{i_1}$. The cost of this rotation can then be added to $\xi_{n_{j_1} n_{i_1}}$.

The complete in-hand path search algorithm is summarized in Algorithm~\ref{algorithm_in-hand_search}. The initial check for the start and goal configurations belonging to the same DMG component ensures that the process of graph search is not initiated when there is certainty of non existence. However, depending on the object's shape, there can be cases in which the condition in (\ref{eq_component_intersection}) is not verified but the in-hand path does not exist nonetheless. In this case, since Dijkstra's algorithm is executed for the nodes in the component $C_1$ and not for all the nodes in the graph, the path is assumed non existent if, after visiting all the nodes in $C_1$, there is no path that has a distance shorter than ${+}\infty$.

\begin{algorithm}[t]
	\small
	
	\SetNoFillComment
	\DontPrintSemicolon
	\SetKwInOut{Input}{Input}
	\SetKwInOut{Output}{Output}
	
	\Input{DMG, $P$, $c_{s_1}$, $c_{s_2}$, $c_{d_1}$, $c_{d_2}$}
	$n_{s_1}, n_{s_2}, n_{d_1}, n_{d_2}\leftarrow$ nodes corresponding to $c_{s_1}, c_{s_2}, c_{d_1}, c_{d_2}$\\
	$C_1\leftarrow$DMG component containing $n_{s_1}$\\
	$C_2\leftarrow$DMG component containing $n_{s_2}$\\
	\If{$n_{d_1}\notin C_1$ \upshape{or} $n_{d_2}\notin C_2$}{
		\Return Null \tcp*{there is no in-hand path}
		}	
	$Q\leftarrow C_1$\\
	\ForEach{$n\in C_1$}{
		dist[$n$]$\leftarrow{+}\infty$\\
		prev[$n$]$\leftarrow$undefined\\
		}
	dist[$n_{s_1}$]$\leftarrow 0$\\
	\While{\upshape{$Q$ is not empty}}{
		$n_{j_1}\leftarrow$ node in $Q$ with minimum \textbf{dist}[$n_j$]\\
		\If{$n_{j_1}\neq n_{d_1}$}{
			\ForEach{$n_{i_1}\in C_1$ \upshape{connected to} $n_{j_1}$}{
				$I\leftarrow$points of intersection between $P$ and the ray from $\textbf{p}_{i_1}$ directed as ${-}\hat{\textbf{n}}_{i_1}$\\
				valid\_node$\leftarrow$false\\
				\ForEach{\upshape{$\textbf{p}\in I$}}{
					$N_{i_2}\leftarrow$ set of nodes with $\textbf{p}$ as centroid\\
					\ForEach{$n_{i_2}\in N_{i_2}$}{
						\If{$n_{i_2}\in C_2$}{
							$A_{i_2}'\leftarrow A_{i_2}$ in $C_1$ reference frame\\
							\If{$A_{i_1}\cap A_{i_2}'\neq\emptyset$}{
								replace $n_{i_1}$ with $\langle \textbf{p}_{i_1}, A_{i_1}\cap A_{i_2}' \rangle$\\
								valid\_node$\leftarrow$true\\
								\textbf{break}
								}
							}
						}
					}
				\If{\upshape{valid\_node}}{
					new\_dist$\leftarrow$dist[$n_{j_1}$] + $\xi_{n_{j_1}n_{i_1}}$\\
					\If{\upshape{new\_dist $<$ dist[$n_{i_1}$]}}{
						dist[$n_{i_1}$]$\leftarrow$new\_dist\\
						prev[$n_{i_1}$]$\leftarrow n_{j_1}$\\
						}
					}
				\Else{
					dist[$n_{i_1}$]$\leftarrow {+}\infty$\\
					prev[$n_{i_1}$]$\leftarrow n_{j_1}$\\
				}
			}
		}
		\Else{
			\tcp{a valid path was found}
			$S\leftarrow$empty sequence\\
			$n\leftarrow n_{d_1}$\\
			\If{\upshape{dist[$n$] is ${+}\infty$}}{
				\Return Null\\	
			}
			\While{\upshape{prev[$n$] is defined}}{
				insert $n$ at the beginning of $S$\\
				$n\leftarrow$prev[$n$]\\
				}
			insert $n_{s_1}$ at the beginning of $S$\\
			\Return $S$
			}
		
	}
	
	\Return Null \tcp*{no valid path was found}
	
	\caption{in-hand\_search}\label{algorithm_in-hand_search}
\end{algorithm}

\subsection{Sequence of Angles}\label{sec_sequence_of_angles}

The in-hand path is a sequence of $K$ nodes $n_{path_0}, ..., n_{path_K}$ that corresponds to the motion of the principal finger on the surface of the object. However, the information about the finger's orientation is not contained in a DMG node. Therefore, a proper sequence of angles must be extracted from the path so that the principal finger is moved from the initial configuration $c_{s_1}$ to the desired one $c_{d_1}$.

While moving between two graph nodes $n_{path_k}{=}\langle \textbf{p}_k, A_{k} \rangle$ and $n_{path_{k{+}1}}{=}\langle \textbf{p}_{k{+}1}, A_{k{+}1} \rangle$, a finger moves from the configuration $c_{k}{=}\langle \textbf{p}_k, \phi_k\rangle$ to the configuration $c_{k{+}1}{=}\langle \textbf{p}_{k{+}1}, \phi_{k{+}1}\rangle$. If $\phi_{k}\in A_{k{+}1}$, there is no need to change the current orientation for the finger to transition into the next node, so $\phi_{k{+}1}{=}\phi_k$. Otherwise, the new angle is chosen so that it satisfies $\phi_{k{+}1}\in A_k \cap A_{k{+}1}$.

\begin{figure}
	\centering
	\begin{subfigure}[b]{0.5\columnwidth}
		\centering
		\includegraphics[width=0.88\columnwidth]{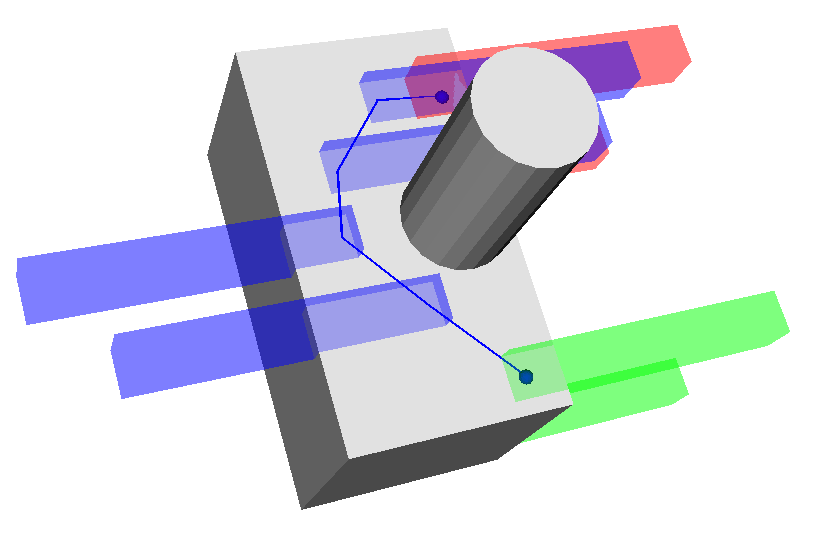}
	\end{subfigure}%
	\begin{subfigure}[t]{0.5\columnwidth}
		\centering
		\includegraphics[width=0.8\columnwidth]{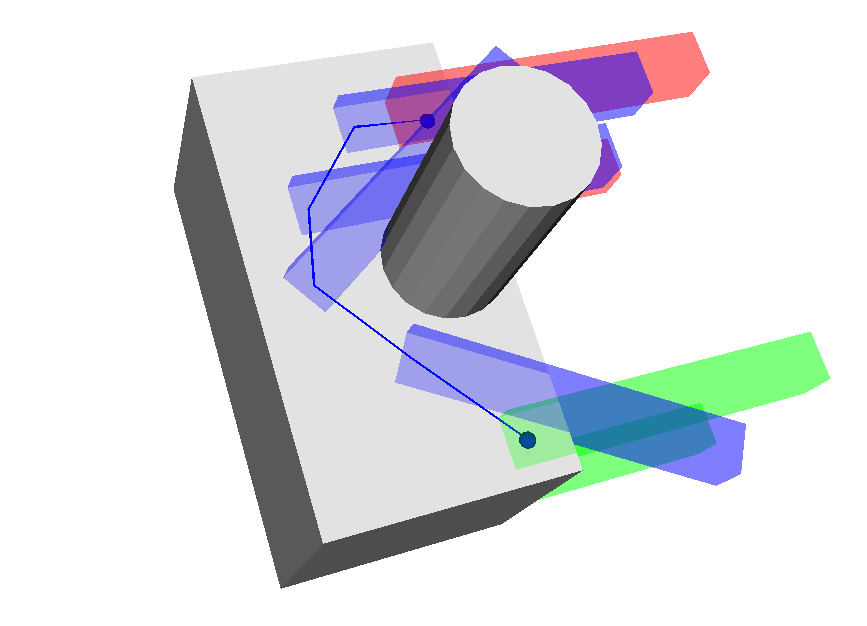}
	\end{subfigure}%
	\caption{Two possible sequences of angles to go from the initial grasp (red) to desired one (green). The fingers in blue show the finger's orientation along the path chosen for the fingertip. In both cases the principal finger rotates counterclockwise around the fingertip contact. In the left image  there is a minimum amount of rotations, in which the finger rotates once to avoid colliding with the object and once more at the end to achieve the desired orientation. In the right image, instead, the finger's angle is kept as close as possible to the desired one, but this results in more rotations: the finger rotates slightly at first, but then it has to rotate once more, always counter-clockwise, to avoid collision, and finally it rotates to obtain the desired angle.}\label{fig_angle_sequence_examples}
\end{figure}

Additional constraints can be added to this angle. For instance, $\phi_{k{+}1}$ can be chosen as the closest one to $\phi_k$ to minimize the immediate rotation. As an alternative, the angle $\phi_{k{+}1}$ can be chosen to minimize the distance from the desired angle or to minimize the total amount of rotations. Fig.~\ref{fig_angle_sequence_examples} shows examples of different possibilities for the rotations. 

In conclusion, to an in-hand path represented as a sequence of nodes we associate a sequence of angles. The final angle is not chosen according to the explained criteria, so it is added last in the sequence and it corresponds to the desired angle. Therefore, the sequence of angles is longer than the sequence of nodes. 

%
%
%
%
%

\subsection{In-Hand Manipulation Sequence}

An in-hand path represented as a sequence of nodes must be transformed into a sequence of in-hand manipulation primitives defined in section~\ref{sec_in-hand_manipulation_primitives}. Therefore, we transform the sequence of nodes in a sequence of translations and rotations.

Due to the construction of the DMG, an edge between two nodes corresponds to a translation. Therefore, from a sequence of $K$ nodes  $n_{path_0}, ..., n_{path_K}$ we can obtain a sequence of $K{-}1$ translations. Given the two consecutive nodes $n_{path_k}$ and $n_{path_{k{+}1}}$, the corresponding translation is defined as the vector 
\begin{equation}
	\textbf{t}_k=\textbf{p}_{{k{+}1}}-\textbf{p}_{k}.
\end{equation}
From the sequence of $K{+}1$ angles corresponding to the sequence of $K$ nodes, as shown in section~\ref{sec_sequence_of_angles}, we can obtain a sequence of $K$ rotations. We assume that a rotation always precedes a translation during the motion execution, and that a final rotation is executed in the end. Given two consecutive angles $\phi_k$ and $\phi_{k{+}1}$, the corresponding rotation is the angle
\begin{equation}
	\gamma_k=\phi_{k{+}1}-\phi_k.
\end{equation}
The sequence of in hand manipulation primitives is an alternation of rotation and translations $\gamma_0, \textbf{t}_0, ..., \gamma_{K{-}1}, \textbf{t}_{K{-}1}, \gamma_K$. This sequence is so that
\begin{equation}
	\textbf{p}_{d_1}=\textbf{p}_{s_1}+\sum_{k{=}0}^{K{-}1}\textbf{t}_k \;\;\mathrm{and}\;\;
	\phi_{d_1}=\phi_{s_1}+\sum_{k{=}0}^{K}\gamma_k.
\end{equation}

\subsection{Sequence Simplification}
Since the DMG is obtained by subdividing the object's shape into small areas, the in-hand manipulation sequence is composed of many segments that connect these areas. However, this sequence can often be simplified because several translations lie on the same line and many rotations are zero. By simplifying the in-hand manipulation sequence, the execution of the motions as described in section~\ref{sec_execution} becomes smoother.

The sequence of translations and rotations is simplified by merging translations that move the fingertip in the same direction when the rotation between them is zero. 

%
%
%
%

\section{Regrasping Planning}
\label{sec_regrasp_planning}
When there is no in-hand path that moves the grasp from the initial fingers' configurations $c_{s_1}$ and $c_{s_2}$ to the desired fingers' configurations $c_{d_1}$ and $c_{d_2}$, the object must be released and regrasped. Depending on the considered system, the regrasping action could also be initiated if the obtained in-hand path is too complex and would result in a much slower execution. This regrasping action completes step 2 in Fig.~\ref{fig_overview}. 

\subsection{Regrasping Sequence Overview}
We take advantage of a dual-arm system so that the object does not have to be properly placed an picked up again, because we use the second gripper as a support. We denote a configuration of a finger of the first gripper, the one initially holding the object, as $c^1$ and a configuration of a finger of the second gripper as $c^2$. Therefore, by first gripper we indicate the gripper for which the desired grasp is defined, and by second gripper we indicate a support gripper.
In the following, we use the superscript to indicate the gripper, while the subscript contains information on the principal or secondary finger.

The goal of the regrasping planning is to:
\begin{itemize}
	\item find release configurations $c_{r_1}^1$ and $c_{r_2}^1$;
	\item find grasp configurations $c_{g_1}^2$ and $c_{g_2}^2$;
	\item find release configurations $c_{r_1}^2$ and $c_{r_2}^2$;
	\item find grasp configurations $c_{g_1}^1$ and $c_{g_2}^1$.
\end{itemize}
A release configuration is the configuration of a finger immediately before the gripper opens, and a grasp configuration is the configuration of the finger immediately after the gripper closes on the object.

A release configuration is needed if the gripper cannot fully open at the current grasp pose due to the object's shape, or if the necessary grasp afterwards is in collision with the current pose of the gripper. For instance, in the example in Fig.~\ref{fig_regrasp2}, the second gripper (blue) is shown after the grasp, but it could slide away from that pose into different release configurations so that the first gripper can directly regrasp in the desired pose. Nonetheless, most of the times for the second gripper the grasp and the release configurations will likely be the same.

The sequence in which the regrasp steps are planned is the following:
\begin{enumerate}
	\item plan a regrasp for the first gripper towards the desired configuration;
	\item find a possible grasp for the second gripper, and plan its release;
	\item plan the release of the first gripper. 
\end{enumerate}
There are two grasps that need to be planned, one per gripper. The criteria with which the grasp is selected differ according to which gripper it is. In fact, the first gripper is the one that must reach the desired configuration, while the second gripper has only a support function. 

\subsection{Antipodal Points}
For grasp planning, we use antipodal point grasps to find a configuration suitable for the two fingers.
Antipodal points are two points on the object's surface whose normals are collinear and in opposite directions. Under the assumption of soft finger contact, these points guarantee force closure~\cite{nguyen_force-closure, mishra_positive_grips}. Instead of looking for antipodal points on the whole object's surface, and since common solutions pose assumptions on the object's shape that the objects we consider do not fulfill~\cite{chen_antipodal_points}, our search constrains the points to be on determined components of the DMG, and in particular we focus only on the centroids of the supervoxels.

Given a point on the object $\textbf{p}_{i_1}$, corresponding to the supervoxel $i_1$, as a candidate contact point for the first fingertip after the grasp, in case there are points intersecting the object along the ray from  $\textbf{p}_{i_1}$ directed as the normal $\hat{\textbf{n}}_{i_1}$, this point is not considered suitable for a direct regrasp as the gripper cannot directly close on it. If it is a suitable candidate, we proceed into evaluating the contact of the second fingertip.

The object intersects the line directed opposite to the normal $\hat{\textbf{n}}_{i_1}$ in one or more points. Since the grasp is achieved by approaching the object with a fully open gripper and then closing the fingers, the selected intersection point is the most external one, i.e. the furthest away from $\textbf{p}_{i_1}$, denoted by $\textbf{p}_{i_2}$. 

Assuming a maximum opening of the gripper equal to $d_{\mathrm{max}}$, these two points are not valid for grasping if
\begin{equation}
	\lVert\textbf{p}_{i_1}-\textbf{p}_{i_2} \rVert_2>d_{\mathrm{max}}.
\end{equation}
Otherwise, we consider the two points good antipodal points if the direction of the normal $\hat{\textbf{n}}_{i_2}$ of the supervoxel that contains the intersection point $\textbf{p}_{i_2}$ is along the same line of $\hat{\textbf{n}}_{i_1}$ and with opposite direction. That is
\begin{equation}\label{eq_antipodal}
 \left\{
	\begin{array}{ll}
	\lVert\hat{\textbf{n}}_{i_1} \times \hat{\textbf{n}}_{i_2}\rVert<\delta_c\\
	~\hat{\textbf{n}}_{i_1}\cdot\hat{\textbf{n}}_{i_2}<0
	\end{array},\right.
\end{equation}
where $\delta_c$ is a tolerance threshold to allow for grasps also in surfaces that are not perfectly parallel. While (\ref{eq_antipodal}) ensures only that the two normals are parallel and opposite, the collinearity is already ensured by the procedure to find $\textbf{p}_{i_2}$.

The choice of fingertip contact point $\textbf{p}_{i_1}$ depends on which grasp is being planned, as the grasp defined by the two antipodal contact is subject to different constraints for the first and for the second gripper. Similarly to the in-hand path solution, a regrasp can differ depending on which finger is chosen as principal finger. 

\subsection{First Gripper Regrasp}\label{sec_first_gripper_regrasp}
\begin{algorithm}[t]
	\small
	
	\SetNoFillComment
	\DontPrintSemicolon
	\SetKwInOut{Input}{Input}
	\SetKwInOut{Output}{Output}
	
	\Input{DMG, $P$, $c_{s_1}$, $c_{s_2}$, $c_{d_1}$, $c_{d_2}$}
	$S^1\gets$ in\_hand\_search(DMG, $P$, $c_{s_1}$, $c_{s_2}$, $c_{d_1}$, $c_{d_2}$)\\
	\If{$S^1\neq$\upshape Null}{
		\tcc{Only one in-hand path is needed, no regrasp}
		\Return Null, Null, $S^1$, Null, Null, Null, Null\\
	}
	\tcc{first gripper regrasp}
	\If{\upshape{can regrasp in $c_{d_1}$, $c_{d_2}$}}{
		$c_{g_1}^1$, $c_{g_2}^1\gets c_{d_1}$, $c_{d_2}$\\
		$S^1\gets$ Null\\	
	}
	\Else{
		\tcc{This iteration starts from $\textbf{n}_{d_1}$ and proceeds backwards from the goal}
		\ForEach{\upshape{$\textbf{p}_{i_1}^1$ centroid of node   $n_{i_1}\in C_{d_1}$}}{
			$\textbf{p}_{i_2}^1\gets$ centroid opposite to $\textbf{p}_{i_1}^1$\\
			\If{\upshape{$\textbf{p}_{i_2}^1$ and $\textbf{p}_{i_1}^1$ are antipodal and can be grasped}}{
				$\phi_{i_1}^1$, $\phi_{i_2}^1\gets$ closest angles to $\phi_{d_1}$\\
				$c_{g_1}^1$, $c_{g_2}^1\gets \langle\textbf{p}_{i_1}^1,\;\phi_{i_1}^1\rangle$, $\langle\textbf{p}_{i_2}^1,\;\phi_{i_2}^1\rangle$\\
				$S^1\gets$ in\_hand\_search(DMG, $P$, $c_{g_1}^1$, $c_{g_2}^1$, $c_{d_1}$, $c_{d_2}$)\\
				\textbf{break}\\
			}
		}
		
	}
	\tcc{second gripper grasp and release}
	$c_{g_1}^2$, $c_{g_2}^2\gets$ regrasp configuration maximizing $d_n$ in (\ref{eq_grasp_line_distance})\\
	\If{\upshape{$d_n$ is $0$}}{
		$c_{g_1}^2$, $c_{g_2}^2\gets$ regrasps far away from $c_{s_1}$, $c_{s_2}$\\
		$c_{g_1}^1$, $c_{g_2}^1\gets$ configurations far away from $c_{g_1}^2$, $c_{g_2}^2$ so that $c_{g_1}^1\in C_{d_1}$ and $c_{g_2}^1\in C_{d_2}$\\
		$S^1\gets$ in\_hand\_search(DMG, $P$, $c_{g_1}^1$, $c_{g_2}^1$, $c_{d_1}$, $c_{d_2}$)\\
		\If{\upshape{$d_n > \epsilon$}}{
			$c_{r_1}^2$, $c_{r_2}^2\gets c_{g_1}^2$, $c_{g_2}^2$\\
			$S^2\gets$ Null\\
		}
		\Else{
			$c_{r_1}^2$, $c_{r_2}^2\gets$ configurations far away from $c_{g_1}^1$, $c_{g_2}^1$, so that $c_{r_1}^2\in C_{g_1}^2$ and $c_{r_2}^2\in C_{g_2}^2$ \\
			$S^2\gets$ in\_hand\_search(DMG, $P$, $c_{g_1}^2$, $c_{g_2}^2$, $c_{r_1}^2$, $c_{r_2}^2$)\\	
		}
	}
	\Else{
		$c_{r_1}^2$, $c_{r_2}^2\gets c_{g_1}^2$, $c_{g_2}^2$\\
		$S^2\gets$ Null\\
	}
	\tcc{first gripper release}
	\If{\upshape{$d_n$ is too small, or direct release not possible}}{
		$c_{r_1}^1$, $c_{r_2}^1\gets$ regrasp configuration maximizing $d_n$ in (\ref{eq_grasp_line_distance}), so that $c_{r_1}^1\in C_{s_1}$ and $c_{r_2}^1\in C_{s_2}$\\
		$S^0\gets$ 	in\_hand\_search(DMG, $P$, $c_{s_1}$, $c_{s_2}$, $c_{r_1}^1$, $c_{r_2}^1$)\\
	}
	\Else{
		$c_{r_1}^1$, $c_{r_2}^1\gets c_{s_1}$, $c_{s_2}$\\
		$S^0 \gets$ Null\\
	}
	
	\Return $S^0$, $S^2$, $S^1$, [$c_{r_1}^1$, $c_{r_2}^1$], [$c_{g_1}^2$, $c_{g_2}^2$], [$c_{r_1}^2$, $c_{r_2}^2$], [$c_{g_1}^1$, $c_{g_2}^1$]
	
	\caption{DMG\_search}\label{algorithm_DMG_search}
\end{algorithm}

The regrasp planned for the first gripper takes into account the desired fingers configurations $c_{d_1}$ and $c_{d_2}$. The goal is to plan the two grasps configurations $c_{g_1}^1$ and $c_{g_2}^1$ so that it is easy for the gripper to reach the target pose. If it is possible to directly regrasp at the desired pose, then $c_{g_1}^1{=}c_{d_1}$ and $c_{g_2}^1{=}c_{d_2}$. Otherwise, the grasp configurations are found by looking for feasible antipodal contact points. These points are searched starting from all the points $\textbf{p}_{i_1}^1$ that correspond to the centroid of nodes in the DMG component $C_{d_1}$. $n_{d_1}$ is used as a starting point so that the selected nodes are as close to it as possible. By keeping the followed connections in memory, the subsequent search for in-hand solution can be optimized.

 The antipodal point $\textbf{p}_{i_2}^1$ must belong to a supervoxel that corresponds to a node in the DMG component $C_{d_2}$. The angles $\phi_{i_1}^1$ and $\phi_{i_2}^1$ are chosen among the angles that keep the fingers in the desired DMG components so that the gripper's orientation points towards the first gripper's side of the robot, and then to minimize the difference between $\phi_{i_1}^1$ and $\phi_{d_1}^1$.  The chosen grasp configurations are then given by $c_{g_1}^1{=}\langle \textbf{p}_{i_1}^1, \phi_{i_1}^1\rangle$ and $c_{g_2}^1{=}\langle \textbf{p}_{i_2}^1, \phi_{i_2}^1\rangle$.
 
It is to be expected that the execution of the grasp will not lead the fingertips in contact exactly at the two chosen points; however, the contacts will belong to components in the DMG from which it is easy to manipulate the object as desired.

\subsection{Second Gripper Grasp and Release}
The second gripper is chosen to maximize the distance from both the current first gripper's grasp and the planned first gripper's regrasp. 

Since the second gripper has only the function of support, there is no constraint on which DMG components the two antipodal points $\textbf{p}_{i_1}^2$ and $\textbf{p}_{i_2}^2$ should belong to. Their search is biased towards points that are as far away as possible from the first gripper's grasp and regrasp points.

First, each node in the DMG is weighted according to the distance between its centroid an the grasp line (i.e. the line between the two fingertips) of the first gripper. More specifically, given the distance from two regrasp lines of the first gripper, one for the current grasp $d_{grasp}$ and one for the planned regrasp $d_{regrasp}$, we associate two values $d_{sum}{=}d_{grasp}{+}d_{regrasp}$ and $d_{diff}{=}|d_{grasp}-d_{regrasp}|$. To each node, the distance value is chosen as
\begin{equation}\label{eq_grasp_line_distance}
	d_{n}=\max(0,\; d_{sum}{-}\zeta d_{diff}),
\end{equation}
in which $\zeta$ is a weighting coefficient. Candidate nodes for the second gripper grasp are those with highest $d_n$. The angles $\phi_{i_1}^2$ and $\phi_{i_2}^2$ are chosen, given the nodes, as angles that allow the fingers to point towards the second gripper's side of the robot and maximize the distance between the first gripper's release and regrasp configurations. 
The release configurations of the second gripper, in this case, coincide with the grasp configurations, i.e. $c_{r_1}^2{=}c_{g_1}^2{=}\langle \textbf{p}_{i_1}^2, \phi_{i_1}^2\rangle$ and $c_{r_2}^2{=}c_{g_2}^2{=}\langle \textbf{p}_{i_2}^2, \phi_{i_2}^2\rangle$.

\begin{figure}
	\centering
	\includegraphics[width=0.7\columnwidth]{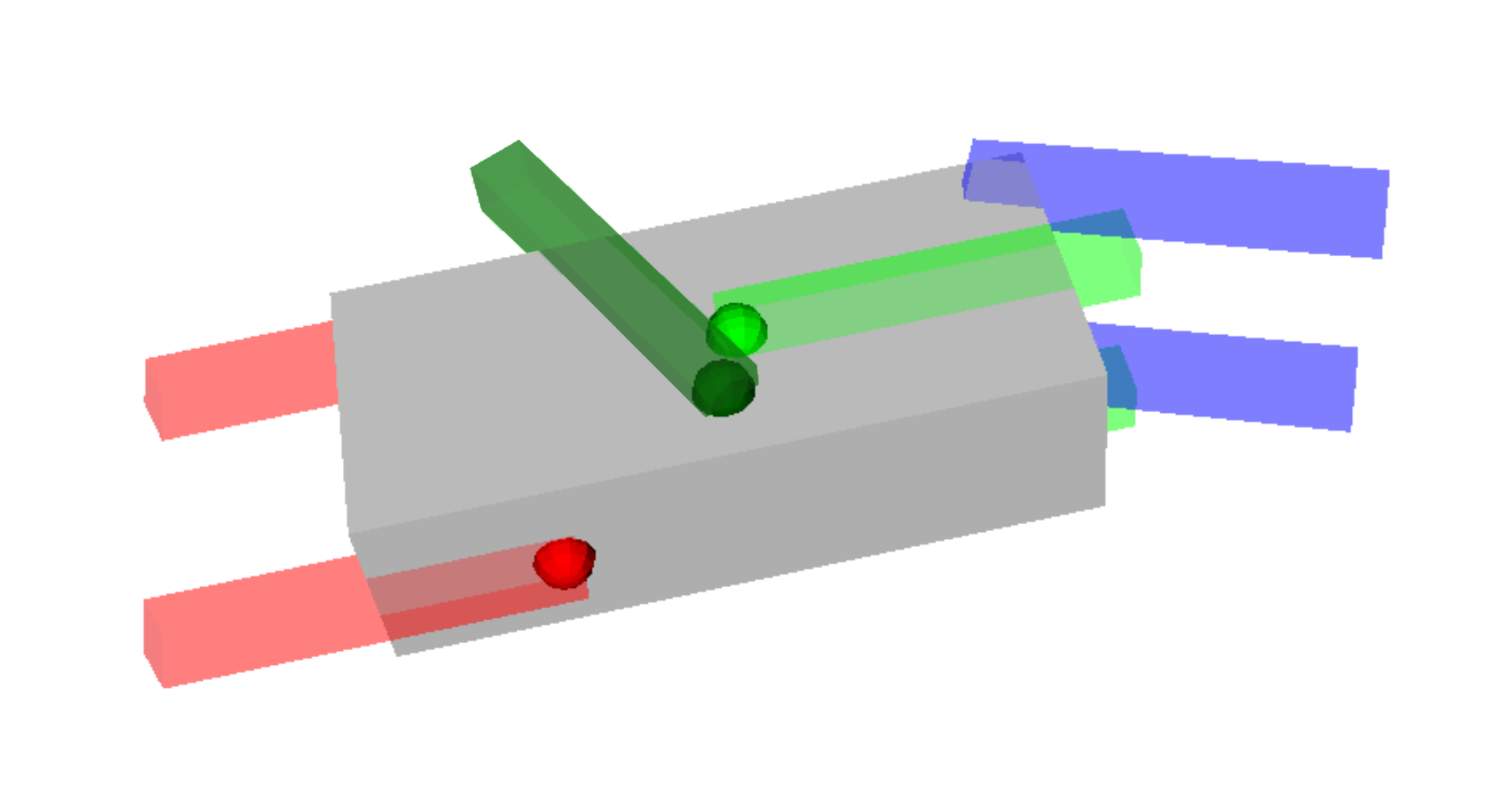}
	\caption{In this example, the current grasp (red) and the desired grasp (light green) hinder feasible grasps for the second gripper. The solution moves the regrasp of the first gripper to the configuration shown in dark green, from which the desired configuration can be achieved through in-hand manipulation, so that the second gripper can grasp as shown by the blue fingers.}\label{fig_regrasping_simple_example}
\end{figure}

If there are no suitable candidate nodes, i.e. $d_n{=}0$ for all the nodes, a new search is performed preferring distance only from the current initial grasp. The two configurations $c_{g_1}^1$ and $c_{g_2}^1$ previously computed as in section~\ref{sec_first_gripper_regrasp} should be modified to increase the distance from the newly found $c_{g_1}^2$ and $c_{g_2}^2$. An example in which the first gripper regrasp configuration is changed is shown in Fig.~\ref{fig_regrasping_simple_example}. If the distance cannot be increased sufficiently, then the second gripper must proceed into the release configurations $c_{r_1}^2$ and $c_{r_2}^2$ that are found to to be far away from the planned regrasp of the first gripper. The search for the grasp and release configurations is subject to keep them in the same DMG components $C_{g_1}^2$ and $C_{g_2}^2$, so that an in-hand path between them can be found. If it is still not possible to find configurations that allow the distance between the grippers to keep a tolerable threshold, the next step is to move away the first gripper and plan according to the new pose, as explained in the next session. However, we do not expect this situation to be frequent.

\subsection{First Gripper Release}

Most of the times $c_{r_1}^1$ and $c_{r_2}^1$ coincide with $c_{s_1}$ and $c_{s_2}$. However, in case this gripper is required to be moved, it is done along the DMG components in which the fingers currently lie. The new fingers's configurations are chosen among the nodes that lie in $C_{s_1}$ and $C_{s_2}$ so that they increase the distance from the second gripper grasp configurations $c_{g_1}^2$ and $c_{g_2}^2$.

\subsection{Summary}

Algorithm~\ref{algorithm_DMG_search} summarizes how to use the DMG with both the in-hand manipulation search and the regrasp planning. 
The obtained solution consists of the regrasp and release configurations for the two grippers and three in-hand sequences. The sequence $S^0$ moves the first gripper from the initial configuration $[c_{s_1}, c_{s_2}]$ to the release configuration $[c_{r_1}^1, c_{r_2}^1]$; the sequence $S^1$ moves the second gripper from the grasp configuration $[c_{g_1}^2, c_{g_2}^2]$ to the release configuration $[c_{r_1}^2, c_{r_2}^2]$; finally, the sequence $S^1$ moves the first gripper from the grasp configuration $[c_{g_1}^1, c_{g_2}^1]$ to the desired configuration $[c_{d_1}, c_{d_2}]$.

We defined a procedure for regrasp that focuses on moving the gripper that is initially grasping the object towards the desired grasp pose. However, if the desired grasp does not pose constraints on which gripper has to be used, our procedure can be simplified because there is only one regrasp involved. More specifically, there is no need for the first gripper regrasp (section~\ref{sec_first_gripper_regrasp}), and the search for the second gripper grasp should be constrained to $C_{d_1}$, $C_{d_2}$ and the distance $d_n$ should try to minimize the distance from the goal while keeping the grasp far away from the first gripper.


\section{Execution}
\label{sec_execution}
Once an object repositioning has been planned, we use a dual-arm robot to perform the execution. In this section we briefly summarize the ECTS formulation that we use to control the robot, and then we define the the motion procedure for executing the in-hand manipulation primitives and regrasping. This section corresponds to step 3 in Fig.~\ref{fig_overview}.

\subsection{Dual Arm Coordination}\label{sec_dual_arm_coordination}

The ECTS \cite{park_ECTS} is a formulation that enables the control of two robot's arms to achieve a common task. This task is specified by an absolute and a relative motion between the end-effectors. These motions are described by the absolute velocity vector $\dot{\textbf{x}}_a{=}(\textbf{v}_a\; \bm{\omega}_a)^T$ and the relative velocity vector $\dot{\textbf{x}}_r{=}(\textbf{v}_r\; \bm{\omega}_r)^T$, which are 6-dimensional vectors of Cartesian and angular velocities $\textbf{v}$ and $\bm{\omega}$.

The ECTS allows a dual-arm robot to share the task between the two arms with a desired degree of coordination. Given the desired relative and absolute motions, the two arms' velocities $\dot{\textbf{x}}_1$ and $\dot{\textbf{x}}_2$ define how each end-effector will move so that the given coordinated task is achieved.
With $\dot{\textbf{x}}{=}(\dot{\textbf{x}}_1\;\dot{\textbf{x}}_2)^T$ as the vector containing the two end-effector's velocities, and $\dot{\textbf{x}}_E{=}(\dot{\textbf{x}}_a\;\dot{\textbf{x}}_r)^T$ as the vector containing the desired motions, the relation between the two is given by
\begin{equation}
	\dot{\textbf{x}}_E=\left(
	\begin{array}{cc}
	\alpha \textbf{I}_6 & (1-\alpha)\textbf{I}_6\\
	-\textbf{I}_6 & \textbf{I}_6
	\end{array} 	
	\right) \dot{\textbf{x}},
\end{equation}
where $\textbf{I}_6$ is the 6-dimensional identity matrix.
The coefficient $\alpha\in[0,\;1]$ manages the share of a coordinated task between the two arms; for instance, if $\alpha$ is $0.5$ the task is shared evenly and the end-effector's motion is symmetric.

The in-hand manipulation primitives define the motion of the object inside one of the robot's gripper. We describe this motion as relative motion between the two grippers, as one gripper is holding the object and the other one is pushing it to help the in-hand motion. Therefore, we can specify an in-hand manipulation sequence by using $\dot{\textbf{x}}_r$.

The velocity $\dot{\textbf{x}}_a$ specifies an absolute motion of the two grippers. In our experiments we keep it to $0$ to minimize noise in sensors and errors in actuations, but it can be exploited to move the arms while the in-hand manipulation task is executed. We see this possibility useful to accelerate the task execution, for instance, in assembly lines, in which the robot picks up an object, then properly repositions it inside the gripper and then it is supposed to place or insert the object in a certain way in a different part of the workspace.

\subsection{Comments on Single Arm Motions}\label{sec_single_arm_motion}

While we specifically target dual-arm systems as ideal platforms for our method, certain solutions can be implemented also on a single arm system. However, this removes the ability of regrasping and it lowers the object repositioning possibilities. A single arm can execute the in-hand manipulation primitives described in section~\ref{sec_in-hand_manipulation_primitives} by exploiting contacts with the environment, similarly to the works in \cite{chavan-dafle_sampling-based_planner,chavan-dafle_in-hand_manipulation_motion_cones}. The formulation that we propose, based on the ECTS framework, can be kept for a single arm by assuming $\alpha{=}0$, so that only one arm moves to achieve the desired relative motion, and the environment is seen as an immobile second arm.

\subsection{Common Procedure for In-Hand Primitives}\label{sec_procedure_for_in-hand_primitives}

The two in-hand primitives, translation and rotation, both require an external push for the object to move within the gripper. In our case, we use the second robot's gripper to push the object.
The first robot's gripper is holding the object at a certain configuration $c$; the in-hand manipulation primitive has the purpose of changing the grasp on the object to a new configuration $c'$, as explained in section~\ref{sec_in-hand_manipulation_primitives}.
The execution procedure is the following:
\subsubsection{Find push point} 
\begin{figure}[t]
	\centering
	\begin{subfigure}[t]{0.5\columnwidth}
		\centering
		\includegraphics[width=0.95\textwidth]{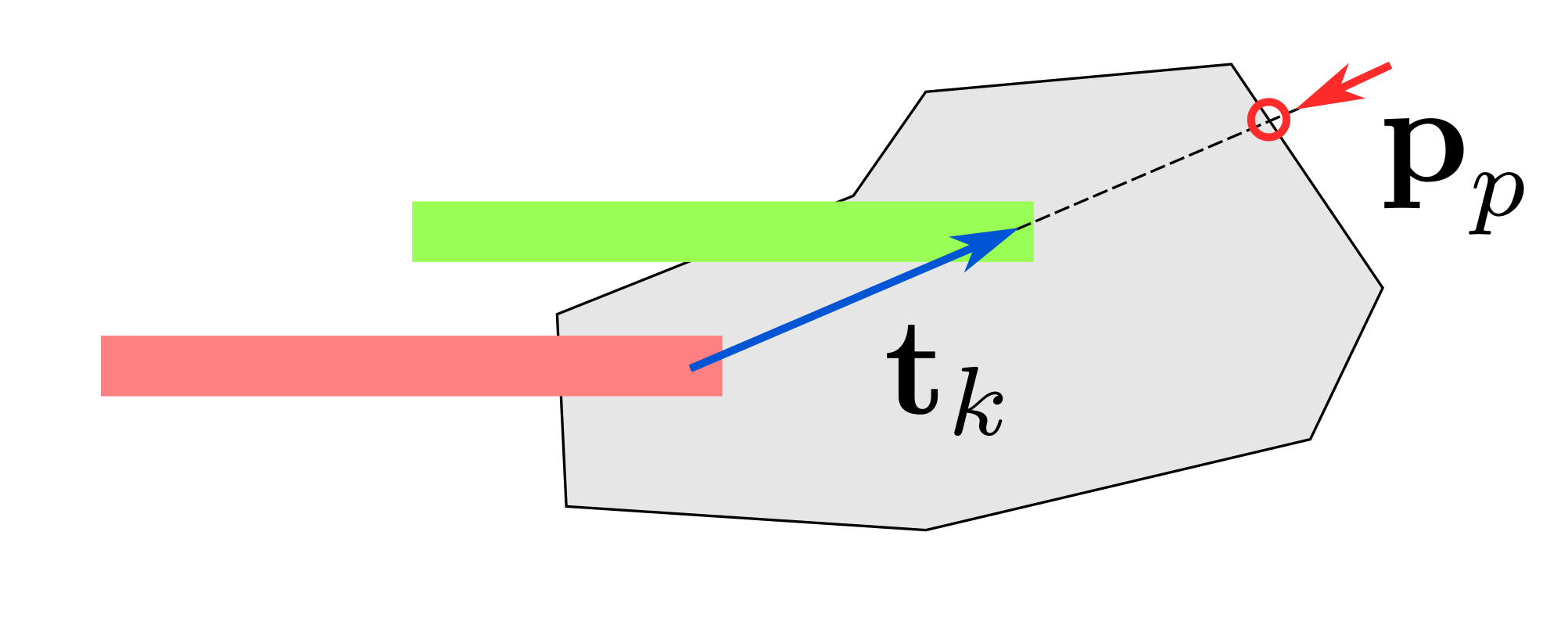}
		\caption{Push point for translation}\label{fig_push_point_translation}
	\end{subfigure}%
	\begin{subfigure}[t]{0.5\columnwidth}
		\centering
		\includegraphics[width=0.95\textwidth]{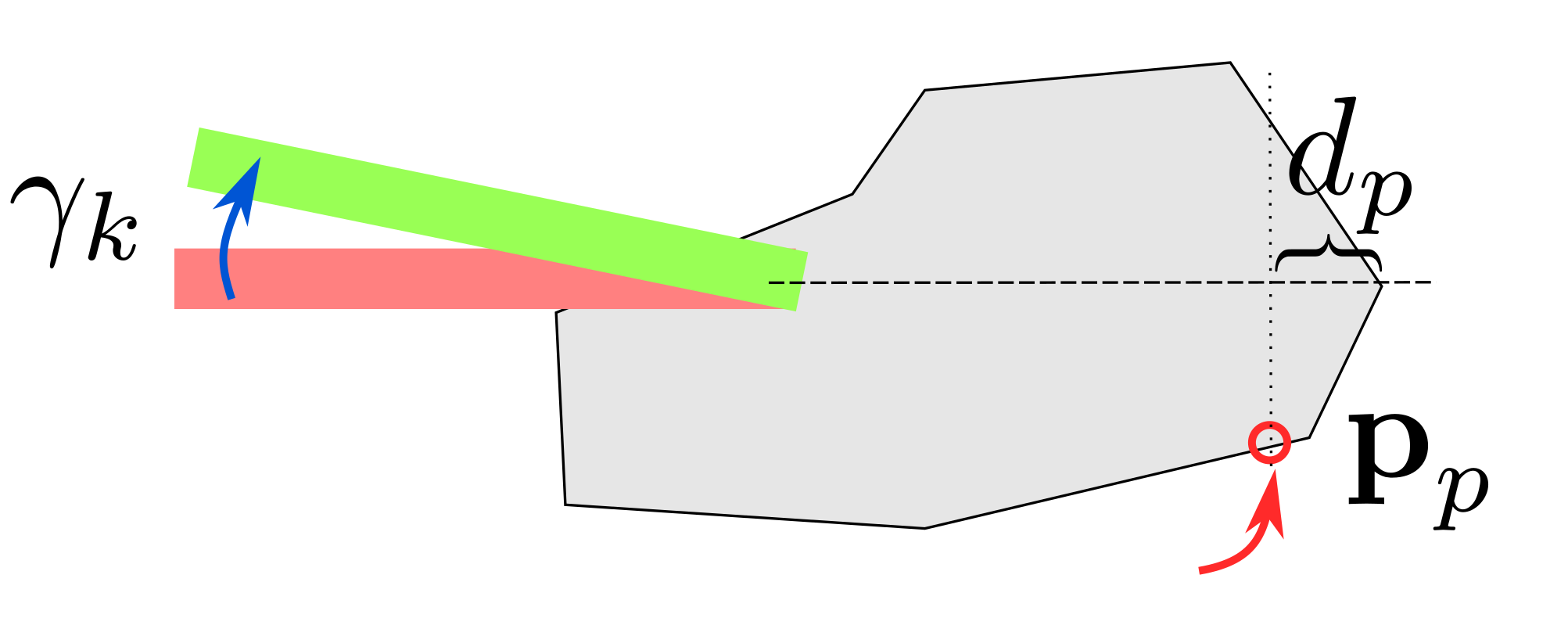}
		\caption{Push point for rotation}\label{fig_push_point_rotation}
	\end{subfigure}
	\caption{The push point is found according to the specific in-hand primitive, translation or rotation. The initial finger's configuration is shown in red and the desired one in green.
	Given a desired translation $\textbf{t}_k$, the push point $\textbf{p}_p$ is found as the intersection between the object's surface and a line starting from the current grasp line along the translation direction. The object is pushed as shown by the red arrow.		
	Given a desired rotation $\gamma_k$, $\textbf{p}_p$ is one of the intersections between the object's surface and a line orthogonal to the finger's direction, moved inside the object by a distance $d_p$. Which of the existing intersections to choose depends on the sign of $\gamma_k$.}\label{fig_push_points}
\end{figure}

A push point is a point on the object from which the second gripper can push so that the motion corresponds to the desired translation or rotation. The procedure to find this contact point depends on the type of motion primitive, and its location depends on the current configuration $c$. Examples of how to find the push point are shown in Fig.~\ref{fig_push_points}. Push point locations are a factor that can affect the feasibility of certain in-hand motions, and they can be taken into account when searching the DMG by altering the cost of edges, as explained in section~\ref{sec_search_for_in-hand_path}.
\subsubsection{Approach push point} The two robot's arms move so that the second gripper is in contact with the object at the push point. Normally only the second gripper's arm is required to move, but both arms can move and adjust the relative pose if necessary, depending on the chosen $\alpha$ value, to improve maneuverability.
\subsubsection{Push} The object is pushed inside the gripper to follow the desired in-hand manipulation primitive. This push is the result of a relative motion between the two arms, so only one or both of them can move according to the desired degree of coordination $\alpha$, as mentioned in section~\ref{sec_dual_arm_coordination}. While the object is pushed, the first gripper enlarges or tightens the grasp to ease the translational or rotational motion.
\subsubsection{Leave push point} The second gripper leaves the contact with the object. As for the approach motion, the two arms can move or only one of them, depending on the need. At the end of this motion, the first gripper is still grasping the object, but in a new configuration $c'$.

\subsection{Translation}\label{sec_motion_execution_translation}
When the in-hand manipulation primitive is translation, the relative motion between the two grippers is a linear motion along the translation direction. This motion is specified by the translation vector $\textbf{t}_k$ in the gripper's reference frame.
By denoting with $\textbf{R}_g$ the rotation matrix describing the orientation of the gripper with respect to the base frame, this vector is expressed in global coordinates as $\textbf{t}_k'{=}\textbf{R}_g\textbf{t}_k$.
For the fingertip to slide as desired, the object must be translated in the opposite direction, denoted by $\hat{\textbf{t}}_k$, that is $\hat{\textbf{t}}_k{=}{-}\frac{\textbf{t}_k'}{\lVert\textbf{t}_k'\rVert}$. This direction corresponds to the direction along which the second gripper should push the object.

The motion of the object inside the gripper corresponds to the relative motion between the two grippers. Our strategy consists in pushing the object with a given velocity until the first gripper's fingertip contact reaches the desired point, i.e. the desired translation has been executed. This velocity is $\textbf{v}{=}m\hat{\textbf{t}}_k$. We adjust the velocity's magnitude $m$ using a P controller.
The dual-arm robot's motion is described by
\begin{equation}\label{eq_relative_velocity_translation}
	\dot{\textbf{x}}_r = \left( \begin{array}{c}
	\textbf{v}\\
	\textbf{0}
	\end{array} \right),
\end{equation}
in which the zero vector indicates that there is no angular velocity, as the object's orientation should not change. When visual feedback is available, we adjust $\dot{\textbf{x}}_r$ to correct errors in the object's motion arising during the execution.

It is assumed that the friction between the second gripper and the object is sufficient to push the latter in the desired direction, even when this direction does not match the surface's normal. However, in case of high mismatches, it is possible to steer the in-hand path search by adjusting the cost of the DMG edges as mentioned in section~\ref{sec_search_for_in-hand_path}. Similarly, translations that require a pull instead of a push, i.e. motions that are in the opposite direction with respect to the first gripper's finger, can be penalized as well, to prefer a sequence of rotations and translations that can be executed more easily. These penalizations are dependent from the initial grasp configuration; therefore they do not modify the DMG, solely object dependent, but they modify only the DMG search strategy.

\subsection{Rotation}
When the in-hand manipulation primitive is rotation, the second gripper should push the object along an arc of a circle centered in the current contact point between the first gripper's fingertip and the object. This contact is fixed on the object, i.e. it does not slide. The relative motion between the two gripper should follow this circle. In addition, we introduce angular velocity so that the contact between the object's surface and the second gripper does not change during this rotational motion.

$\gamma_k$ is the desired rotation specified for the first gripper's finger. The push point $\textbf{p}_p$ on the object's surface should rotate around the first gripper's fingertip contact, in the opposite direction of the desired rotation.
Assuming that the desired rotation $\gamma_k$ is an angle in the yz plane of the first gripper's reference frame, the planar circular motion during time is expressed by the vector $(\cos(\phi_0{+}\gamma(t))\;\; \sin(\phi_0{+}\gamma(t)))^T$. $\phi_0$ is the initial angle between the push point $\textbf{p}_p$ and the first gripper's finger. The function $\gamma(t)$ describes the variation of this angle. Assuming the rotation execution between an initial time $t{=}t_0$ and a final time $t{=}t_f$, this function is so that $\gamma(t_0){=}0$ and $\gamma(t_f){=}{-}\gamma_k$.

We express the relative motion in the first gripper's reference frame as
\begin{equation}\label{eq_relative_velocity_rotation}
	^g\dot{\textbf{x}}_r=\left(
	\begin{array}{c}
	0\\ 
	{-}\sin(\phi_0{+}\gamma(t))\dot{\gamma}(t)\\
	\cos(\phi_0{+}\gamma(t))\dot{\gamma}(t)\\
	\dot{\gamma}(t)\\
	0\\
	0\\
	\end{array}
	\right).
\end{equation}
This velocity is then transformed into a relative velocity $\dot{\textbf{x}}_r$ expressed in the base frame.
Similarly to the translation velocity, we adjust the magnitude of the Cartesian velocity during the execution, until the object reaches the desired angle, and the velocity vector is adapted to compensate for errors when visual feedback on the object's motion is available. 

Since large rotations could lead to collisions between the two grippers, they should be penalized similarly to the penalization of different translation directions during the DMG search.
\begin{table*}
	\centering
	\begin{tabular}{|c|c|c|cc|}
		\hline
		\multirow{3}{*}{\textbf{object}} & 
		\multirow{2}{*}{\textbf{bounding box}} &  
		\multicolumn{3}{c|}{\textbf{Time} [s]}\\
		\cline{3-5}
		& & \multirow{2}{*}{$r_{area}$ [mm]}& \multicolumn{2}{c|}{$r_{angle}$ [deg]}\\
		\cline{4-5}
		& [mm$\times$mm$\times$mm]& & \multicolumn{1}{c|}{5} & 20\\
		\hline
		\multirow{3}{*}{\includegraphics[height=1cm]{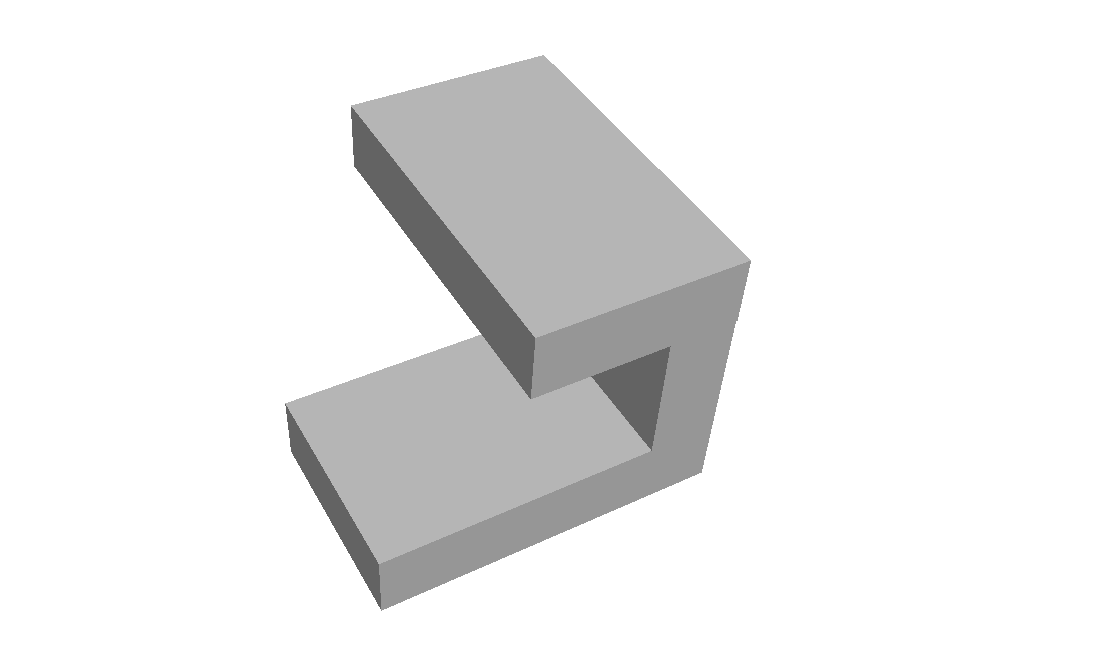}}&\multirow{3}{*}{50$\times$50$\times$40}& 10 &
		1.861 (0.299+1.562) & 1.847 (0.299+1.548)\\
		\cline{3-3}
		& & 15 & 0.945 (0.353+0.592) & 0.943 (0.353+0.590)\\
		\cline{3-3}
		& & 20 & 1.050 (0.465+0.585) & 1.049 (0.465+0.584)\\
		\hline
		\multirow{3}{*}{\includegraphics[height=1cm]{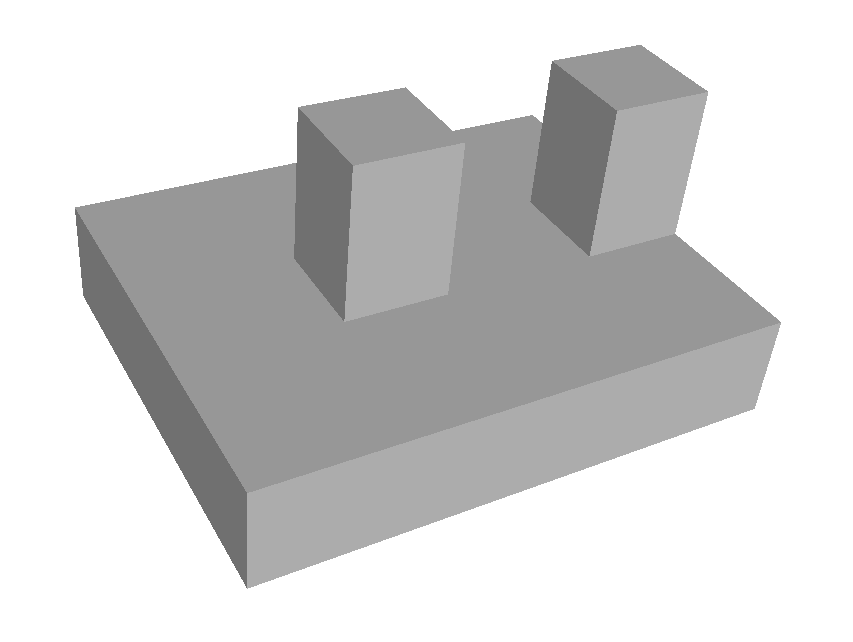}}&\multirow{3}{*}{100$\times$80$\times$50}& 10 &
		11.636 (1.062+10.574) & 11.297 (1.062+10.235)\\
		\cline{3-3}
		& & 15 & 6.953 (1.323+5.630) & 7.066 (1.323+5.743)\\
		\cline{3-3}
		& & 20 & 4.071 (1.412+2.659) & 4.054 (1.412+2.642)\\
		\hline
		\multirow{3}{*}{\includegraphics[height=1cm]{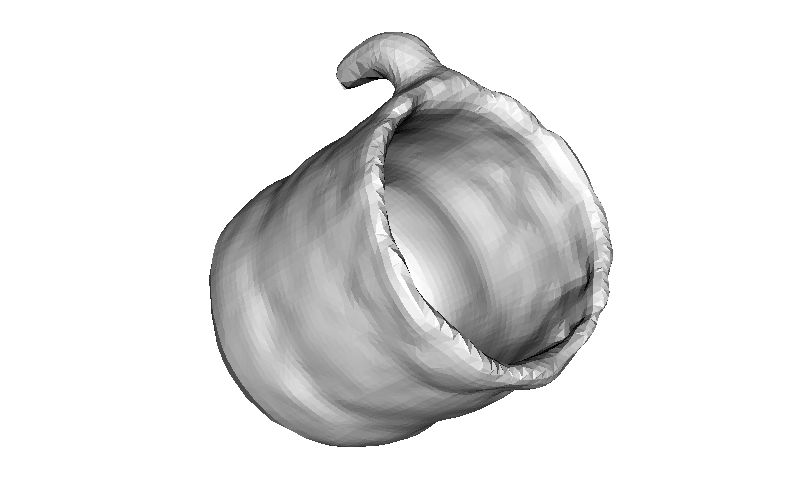}}&\multirow{3}{*}{87.8$\times$107.2$\times$104.8}& 10 & 5.902 (1.302+4.600) & 5.853 (1.302+4.551) \\
		\cline{3-3}
		& & 15 & 3.371 (1.514+1.857) & 3.326 (1.514+1.812)\\
		\cline{3-3}
		& & 20 & 2.738 (1.676+1.062) & 2.738 (1.676+1.062)\\
		\hline
	\end{tabular}
	\caption{Times of DMG generation. The total time is a sum between the time spent in the supervoxel segmentation and the one spent in refining the graph.}
	\label{tab_dmg_generation}
\end{table*}

\subsection{Regrasping}\label{sec_procedure_for_regrasp}
The regrasping execution follows this sequence:
\subsubsection{Reach First Release Configuration} At the beginning, the first gripper is holding the object at the initial fingers' configurations $c^1_{s_1}$ and $c^1_{s_2}$. If the release configurations $c^1_{r_1}$ and $c^1_{r_2}$ do not coincide with them, an initial in-hand motion is performed until they are reached.
\subsubsection{Second Gripper Grasp} The second gripper should grasp the object so that its fingers are in the configurations $c^2_{g_1}$ and $c^2_{g_2}$. If these configurations are difficult to reach given the current relative position of the two robot's arms, the first arm moves so that the line starting from the first gripper to the desired second gripper's grasp points is directed towards the second arm. However, since this side of the robot is preferred during the search for regrasp, the need for this reconfiguration is minimal. Once the desired configurations are within reach, the second gripper closes on the object. At the end of this phase, both grippers are grasping the object.
\subsubsection{First Gripper Release} The first gripper opens the fingers to release the object and it is moved away from it. The object is now held only by the second gripper.
\subsubsection{Reach Second Release Configuration} If the second gripper's release configurations $c^2_{r_1}$ and $c^2_{r_2}$ differ from the grasp configurations, in-hand motion is required to readjust the pose of the object. In this case, the in-hand motion sequence is executed as described previously, with the only difference of assuming the first gripper ar pusher and the second gripper as holding the object.
\subsubsection{First Gripper Grasp} The first gripper should grasp the object to reach the two fingers' configurations $c^1_{g_1}$ and $c^1_{g_2}$. Similarly to the second gripper grasp, the relative pose between the two arms should be properly adjusted when needed. At the end of this phase both grippers are grasping the object.
\subsubsection{Second Gripper Release} The second gripper opens the fingers to release the object and it moves away from it. The object is now held only by the first gripper. If the current grasp configurations are not the desired ones $c^1_{d_1}$ and $c^1_{d_2}$, a final in-hand manipulation execution is necessary to drive the object into the desired pose.

\section{Experiments}
\label{sec_experiments}
In this section we present the results obtained using the DMG for grasp reconfiguration.

\subsection{General Setup}

To obtain the subdivision of an object's shape into small areas, we used the supervoxel method available in the Point Cloud Library (PCL)~\cite{rusu_PCL}. For the robot experiments, we used an ABB Yumi with a Kinect v2 camera mounted on top. The camera was used to detect the object's pose with respect to the gripper's fingertips, using Apriltags~\cite{olson_apriltag}. This pose detected at the beginning of the execution was used to infer the initial fingers configurations, and, during the in-hand manipulation and after regrasp it was used as visual feedback to adjust Yumi's motion.
Yumi's finger implied $l_f$ to be $4$ cm and we used deformable hemispherical fingertips for easing the control of the friction when enlarging and tightening the grasp. These fingertips ensured that the contact with the object could be approximated with a single point.

In our dual-arm experiments, we set $\alpha{=}1$ for the in-hand manipulation part, so that the gripper holding the object would not move and the object's pose detection was easier. We switched to $\alpha{=}0$ when the second gripper was grasping the object.
For the graph refinement, we used a threshold $\delta_n{=}0.15$, and $\delta_c{=}0.4$ for regrasp planning.

\subsection{Graphs Generation} \label{sec_experiments_graph_gen}

A new DMG should be generated for each object we want the robot to manipulate. The complexity of the generation procedure (Algorithm~\ref{algorithm_DMG_generation}) can easily be obtained: with $|N|$ denoting the number of nodes derived from the supervoxels, directly dependent on the chosen resolution $r_{area}$, $K_{neigh}$ as the maximum number of edges per node and $K_{comp}$ as the maximum number of angular component per supervoxel centroid, we can write the complexity as
\begin{equation}\label{eq_complexity}
O((\max((3+K_{comp})K_{neigh}, \;\;2K_{angle}))|N|).
\end{equation}
Since these numbers are subject to geometric constraints of the object's shape, $K_{neigh}$ is usually at most 5 and $K_{comp}$ is 1 for most of the supervoxel centroids, and it can be expected to remain a low number (2 or 3). $K_{angle}$ depends on the chosen resolution for the angles $r_{angle}$. 

We tested the graph generation on different objects, varying the resolution of the supervoxels $r_{area}$ and the angle resolution $r_{angle}$. Table~\ref{tab_dmg_generation} shows the time spent to generate the DMG, which is a sum of the time (in average) required by the supervoxel method to provide a subdivision of the object into small areas and the time to refine the connectivity between the nodes associated to the centroids of these areas. The time has been obtained from running the DMG generation on an Intel Core i7-6700 3.40 GHz processor.  

The main difference between the first and the second object is due to the increase in size. The third object, a mug from the YCB object dataset~\cite{calli_YCB_dataset}, requires less time than the second object to generate the DMG, despite being the biggest object. This is due to the particular object's shape, that presents a curved surface that that causes the mismatch between the normals to exceed the threshold for translation much more frequently. In addition, the object model is noisy. The result is a graph with many small isolated components, for which fewer iterations are required. We increased the value of $\delta_n$ up to $0.3$ and verified that this increase does not produce significant changes both in the graph structure and in the generation time, due to the curvature and the noisy reconstruction of this shape.

These results show that the main change on the graph generation time is given by the different values of $r_{area}$, which has a direct influence on the term $|N|$. The value $r_{angle}$, which instead influences $K_{comp}$ and $K_{angle}$, has a much smaller effect in the variation of the required time.

We think of the DMG as a tool that can be exploited multiple times once it has been generated for a given object, for instance in industrial applications where the same object's shape must be manipulated several times in different ways. However, the times reported in Table~\ref{tab_dmg_generation} show that it is also suitable for single use applications in which a few seconds can be sacrificed before initiating the in-hand manipulation task.

\subsection{Regrasping Execution}

\begin{figure}
	\centering
	\begin{subfigure}[t]{0.59\columnwidth}
		\centering
		\includegraphics[width=0.96\textwidth]{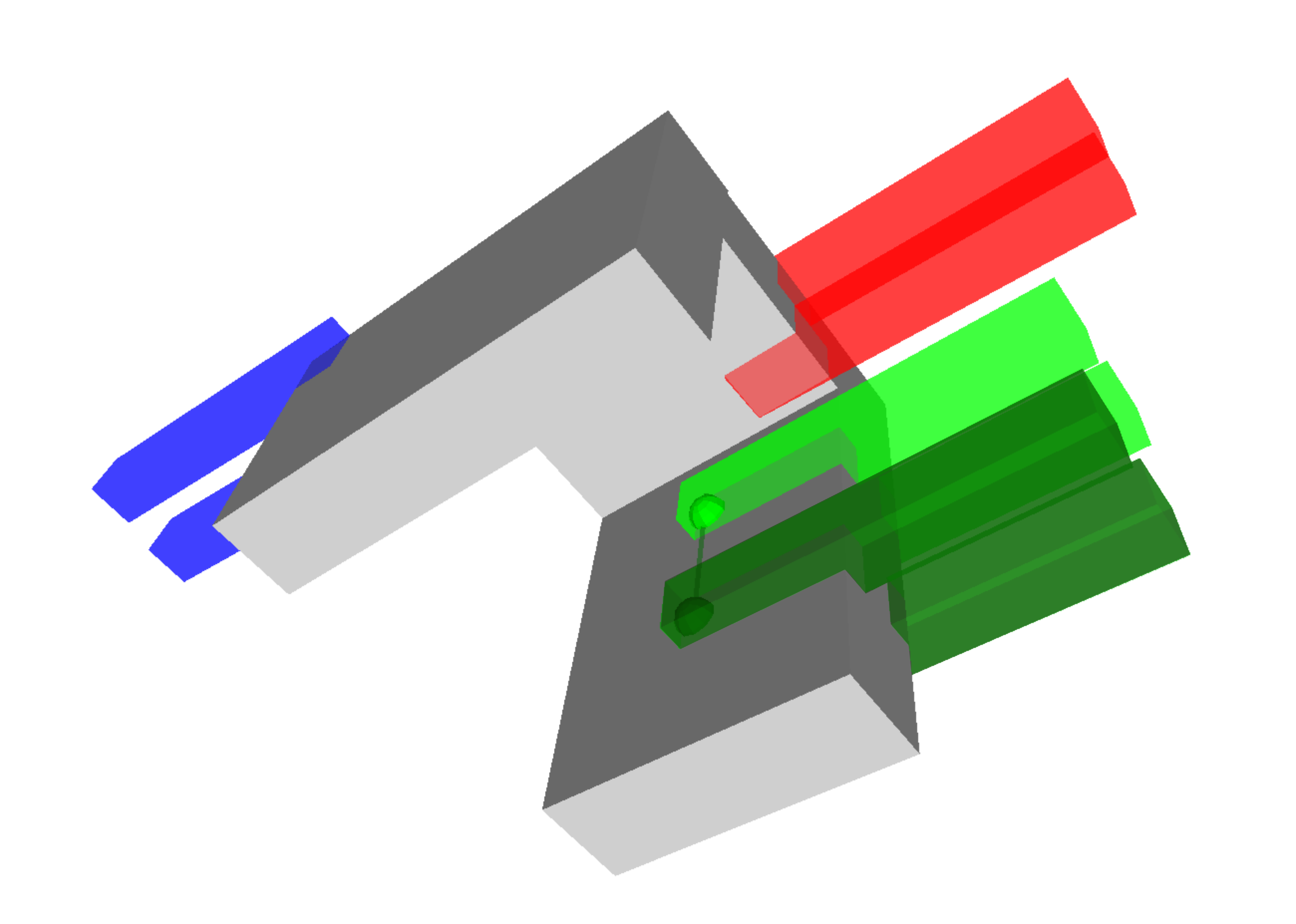}
	\end{subfigure}%
	\begin{subfigure}[t]{0.39\columnwidth}
		\centering
		\includegraphics[width=0.98\textwidth]{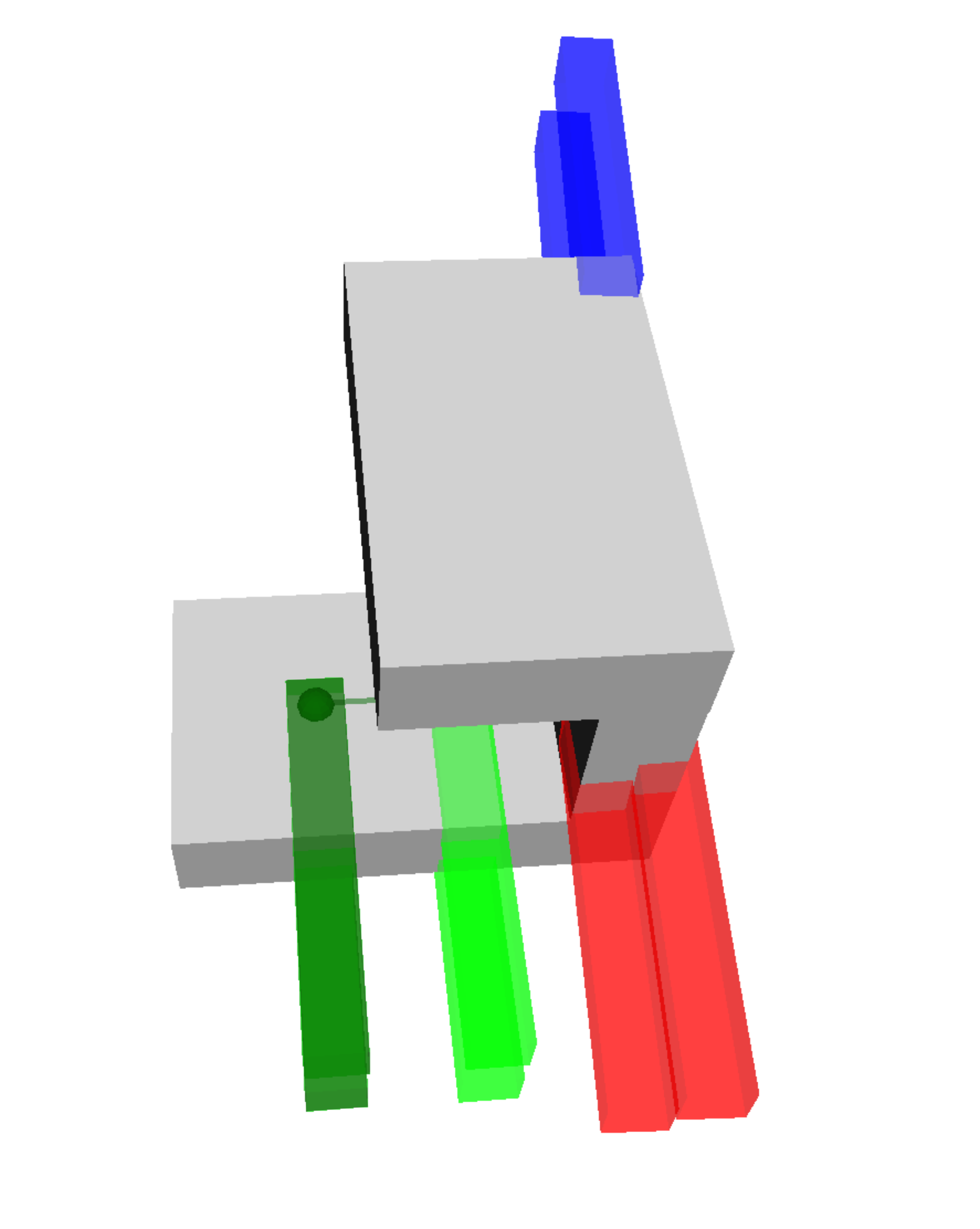}
	\end{subfigure}%
	
	\caption{Two views of the planned grasp reconfiguration. The gripper's fingers should go from the initial configuration, shown in red, to the desired configuration inside the object's concavity, shown in green. The Second gripper should help regrasping by holding the object in the configuration shown by the blue fingers. The first gripper should regrasp the object as shown by the dark green fingers so that the fingertips are in contact as shown by the dark green sphere and then slide inside the concavity to reach the desired configuration.}\label{fig_regrasp_execution_plan}
\end{figure}

\begin{figure}
	\centering
	\begin{subfigure}[b]{\columnwidth}
		\centering
		\includegraphics[width=0.7\textwidth]{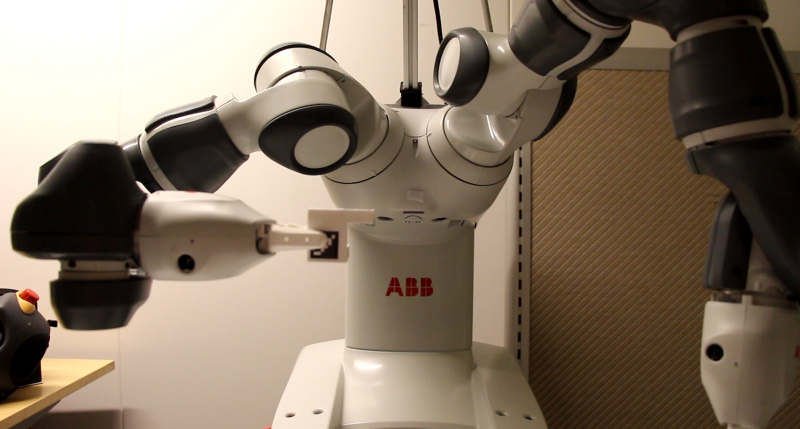}
		\caption{Initial configuration.}\label{fig_regrasp_execution1}
	\end{subfigure}
	\begin{subfigure}[b]{\columnwidth}
		\centering
		\includegraphics[width=0.7\textwidth]{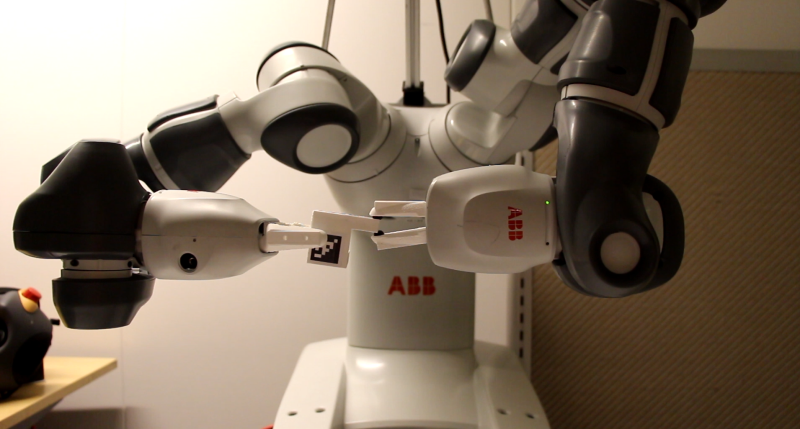}
		\caption{Second Gripper Grasp.}\label{fig_regrasp_execution2}
	\end{subfigure}
	\begin{subfigure}[b]{\columnwidth}
		\centering
		\includegraphics[width=0.7\textwidth]{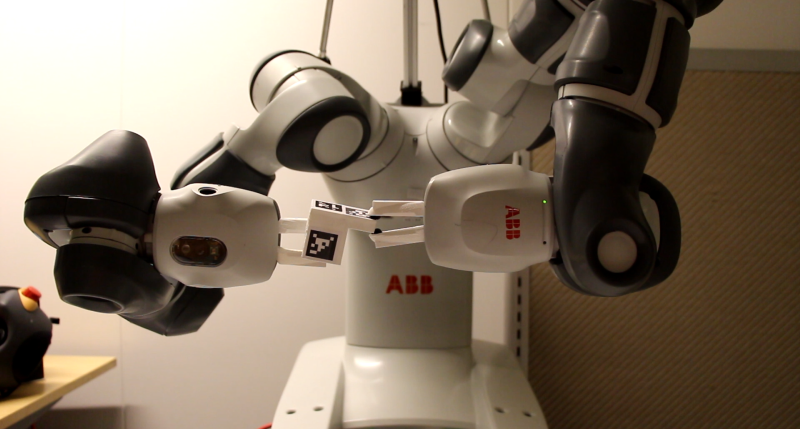}
		\caption{First Gripper Grasp.}\label{fig_regrasp_execution3}
	\end{subfigure}
	\begin{subfigure}[b]{\columnwidth}
		\centering
		\includegraphics[width=0.7\textwidth]{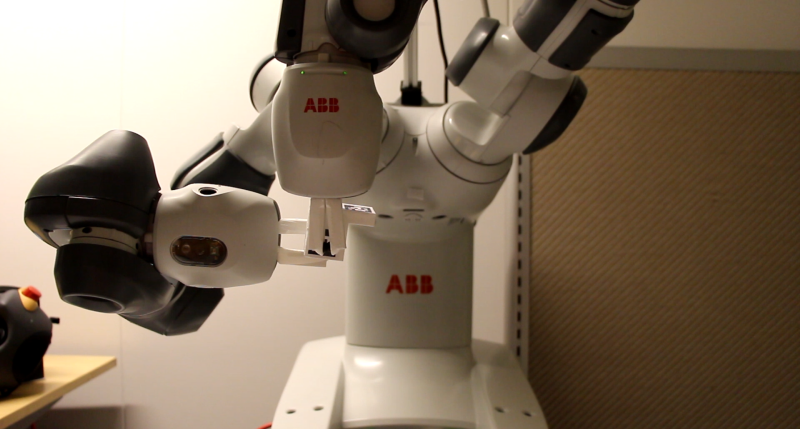}
		\caption{Push.}\label{fig_regrasp_execution4}
	\end{subfigure}
	\begin{subfigure}[b]{\columnwidth}
		\centering
		\includegraphics[width=0.7\textwidth]{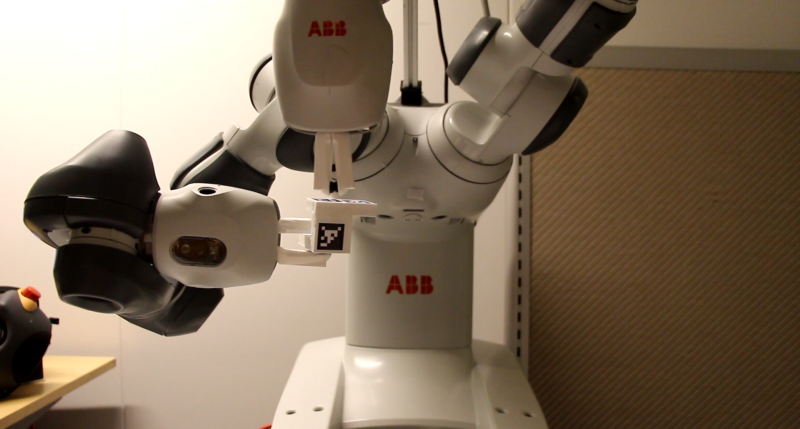}
		\caption{Final configuration.}\label{fig_regrasp_execution5}
	\end{subfigure}
	\begin{subfigure}[b]{\columnwidth}
		\centering
		\includegraphics[width=0.7\textwidth]{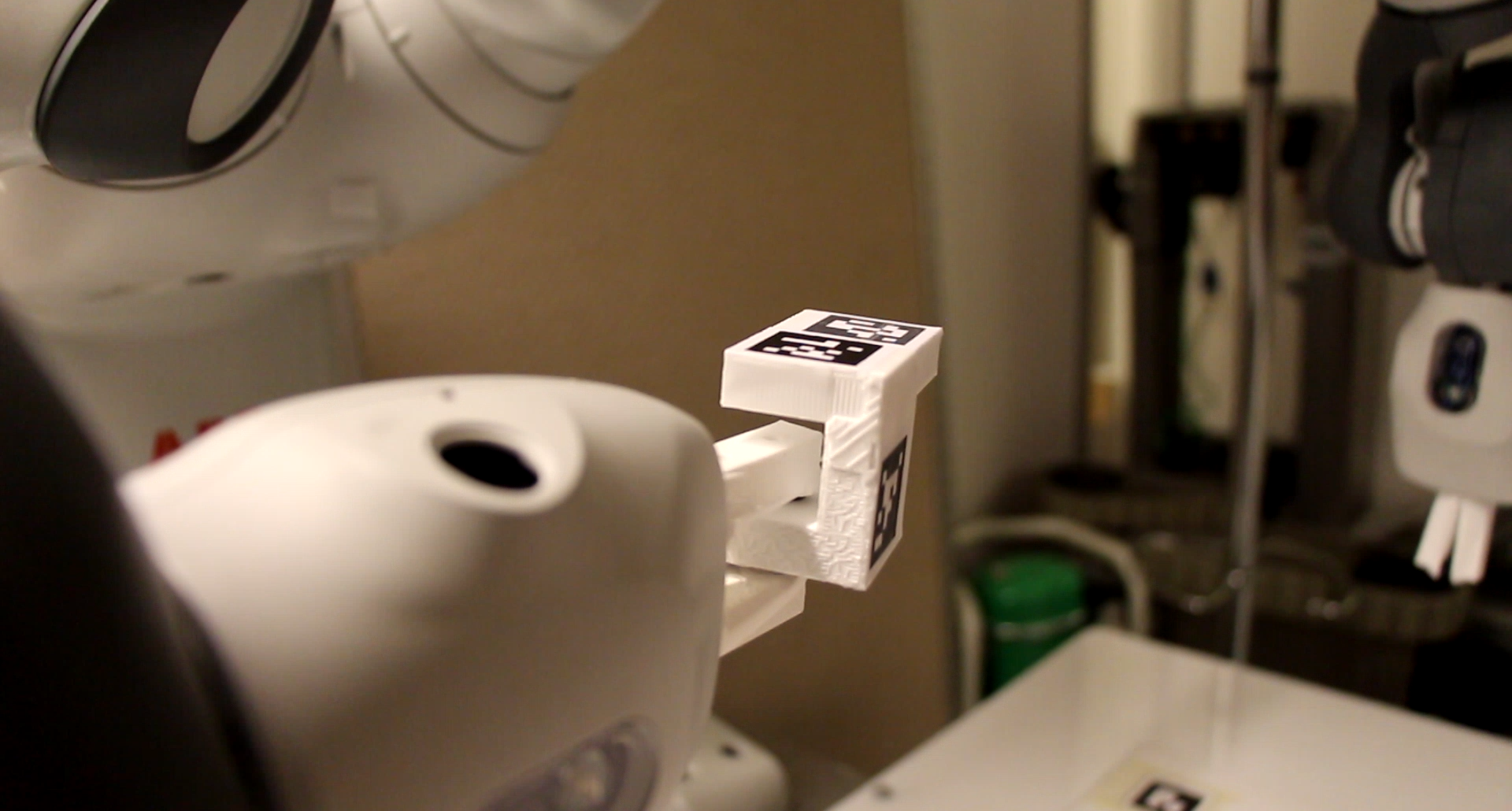}
		\caption{Close view of the final configuration.}\label{fig_regrasp_execution6}
	\end{subfigure}
	\caption{Yumi executing a grasp reconfiguration following the procedures defined in sections~\ref{sec_procedure_for_in-hand_primitives} and~\ref{sec_procedure_for_regrasp}.}\label{fig_regrasp_execution}
\end{figure}

We tested the regrasping strategy combined with in-hand manipulation. We gave the object to one of the robot's grippers and set the goal pose so that regrasping was required to achieve the desired configuration.

In the example shown in Fig.~\ref{fig_regrasp_execution_plan}, the goal pose of the fingers is inside a concavity of the object; directly regrasping the object inside the concavity is not trivial, as the gripper cannot easily approach the desired contact point with open fingers. Therefore, the DMG solution provides a regrasp on the same graph component of the desired configuration, from which the fingertip can slide into the concavity. This solution was obtain from a DMG generated with $r_{area}{=}10$ mm.

The execution of this grasp reconfiguration is shown in Fig.~\ref{fig_regrasp_execution}. 
Yumi is initially grasping the object as shown in Fig.~\ref{fig_regrasp_execution1}; from this initial grasp, the initial configuration of the fingers on the object is obtained, given the object's pose visible from the Kinect. To achieve the desired grasp, Yumi has to regrasp the object, as shown in Figs.~\ref{fig_regrasp_execution2} and \ref{fig_regrasp_execution3}, and then push the object to execute the in-hand manipulation task, as in Fig.~\ref{fig_regrasp_execution4}. After the push, the fingers have reached the desired contact, shown in Fig.~\ref{fig_regrasp_execution5}. 
As it is possible to see from a closer look of the final configuration (Fig.~\ref{fig_regrasp_execution6}), while the contact point is inside the object's concavity as desired, the final pose of the object inside the gripper is slightly rotated with respect to the desired one. This is due to the lack of feedback during the push, as the arm was covering the camera; therefore, the pushing action could not be adjusted during the execution.
However, a new in-hand plan providing a simple rotation action is enough to adjust the final orientation of the object in cases like this.



\begin{figure}
	\begin{subfigure}[t]{0.25\textwidth}
		\centering
		\includegraphics[width=\textwidth]{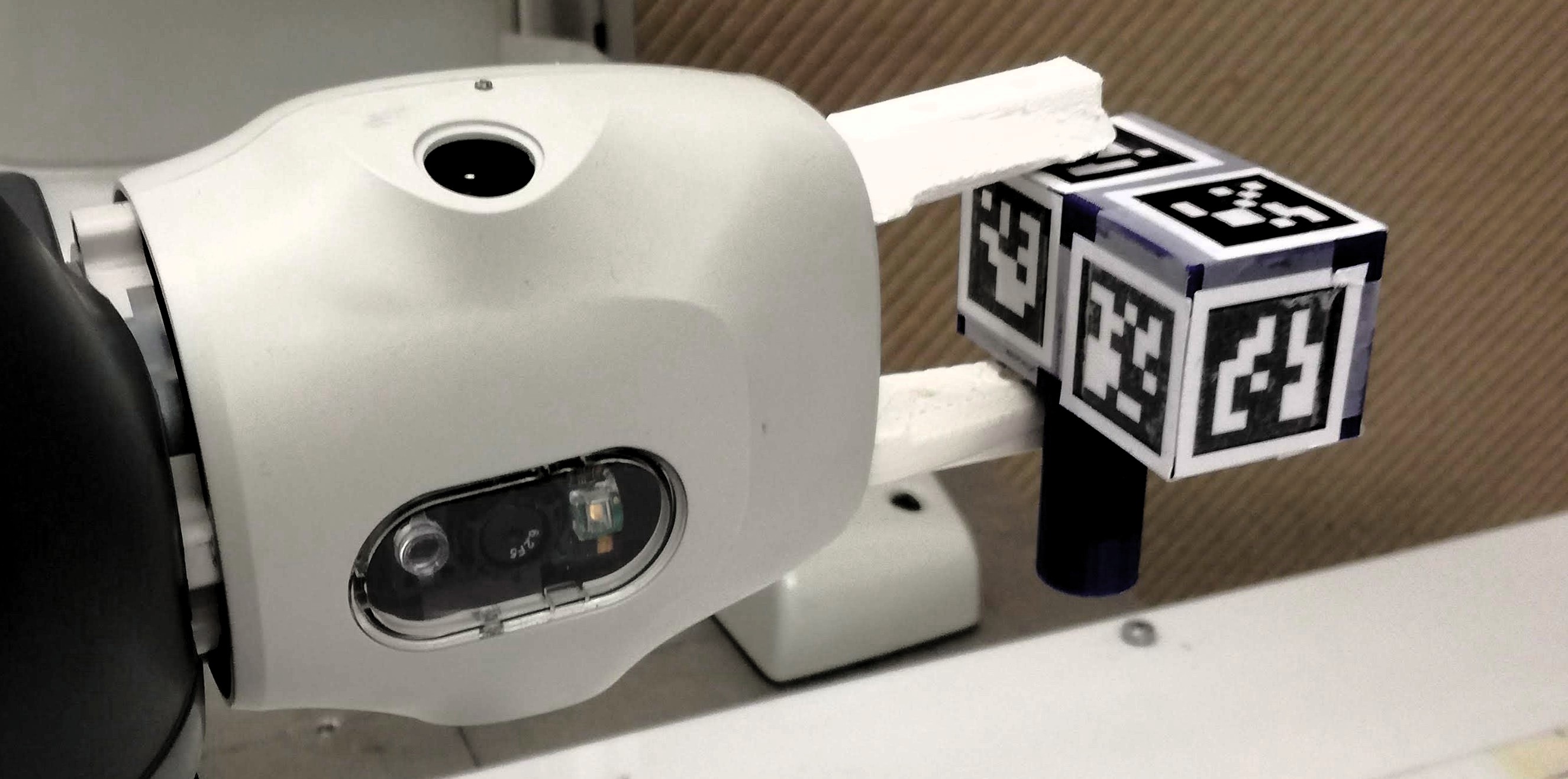}
		\caption{Yumi's initial grasp.}\label{fig_yumi_initial_grasp}
	\end{subfigure}%
	\begin{subfigure}[t]{0.25\textwidth}
		\centering
		\includegraphics[width=\textwidth]{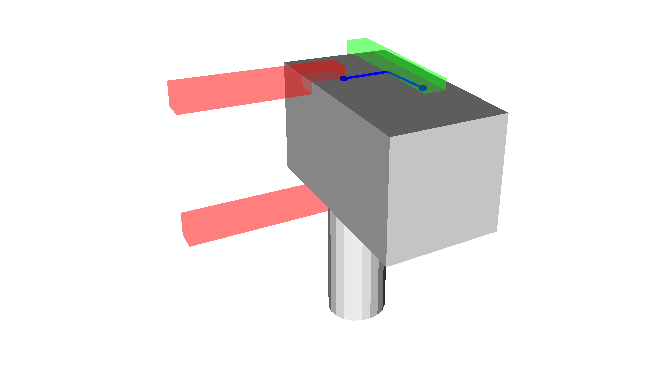}
		\caption{In-hand solution.}\label{fig_in-hand_solution}
	\end{subfigure}
	\caption{An example of in-hand manipulation task that we executed with the Yumi robot. The goal is to move the fingers on the object from the initial configuration to the desired one without releasing the grasp.}
\end{figure}

\begin{figure}
	\centering
	\includegraphics[width=0.4\textwidth]{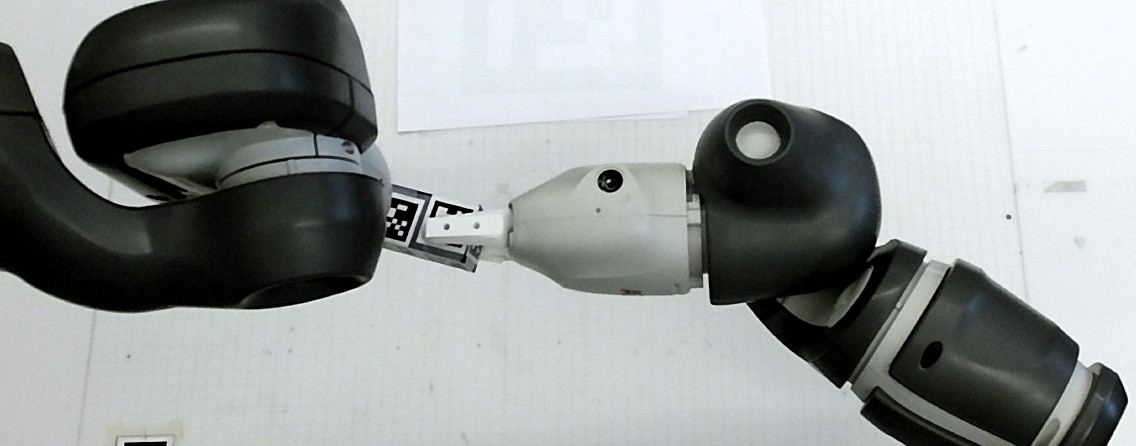}
	\caption{Yumi executing an in-hand rotation by pushing the object with the second gripper along an arc of a circle.}\label{fig_yumi_in-hand_execution}
\end{figure}

\subsection{In-Hand Manipulation Execution}

In our previous work~\cite{cruciani_dexterous_manipulation_graph} we tested different in-hand manipulation solutions using a Baxter robot. We executed more experiments with Yumi and we also show an example of in-hand manipulation exploiting the environment as an immobile second arm.

Fig.~\ref{fig_yumi_initial_grasp} depicts an initial grasp on an object and Fig.~\ref{fig_in-hand_solution} shows the same fingers' configuration on the object, in red, and the desired pose of the fingers, in green. Since the two configurations lie in the same DMG components, i.e. the condition in (\ref{eq_component_intersection}) is not verified, it is possible to move the fingers from one to the others using in-hand manipulation.
The path in blue in Fig.~\ref{fig_in-hand_solution} shows the desired motion of the fingertip along the object's surface. Despite being planned along the flat surface, this solution avoids the obstacle on the opposite side of the object, as explained in section~\ref{sec_search_for_in-hand_path}. The sequence of in-hand motions is composed of two translation segments and a rotation of $90^\circ$. Fig.~\ref{fig_yumi_in-hand_execution} shows Yumi executing the rotational motion of this in-hand motions sequence. 

\begin{figure}
	\centering
	\begin{subfigure}[b]{0.25\textwidth}
	\centering
	\includegraphics[width=\textwidth]{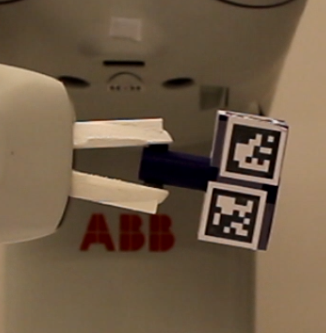}
	\caption{Yumi's initial grasp on the object.}\label{fig_extrinsic_dexterity_initial_grasp}
	\end{subfigure}%
	\begin{subfigure}[b]{0.25\textwidth}
	\centering
	\includegraphics[width=0.7\textwidth]{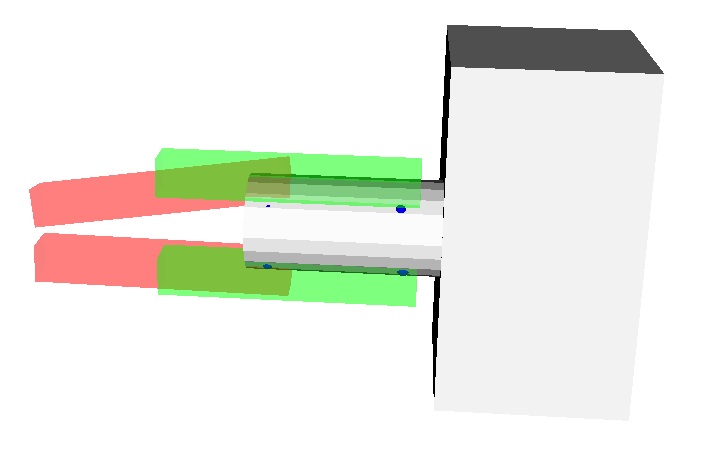}
	\caption{Initial and desired grasp.}\label{fig_extrinsic_dexterity_path}
	\includegraphics[width=0.7\textwidth]{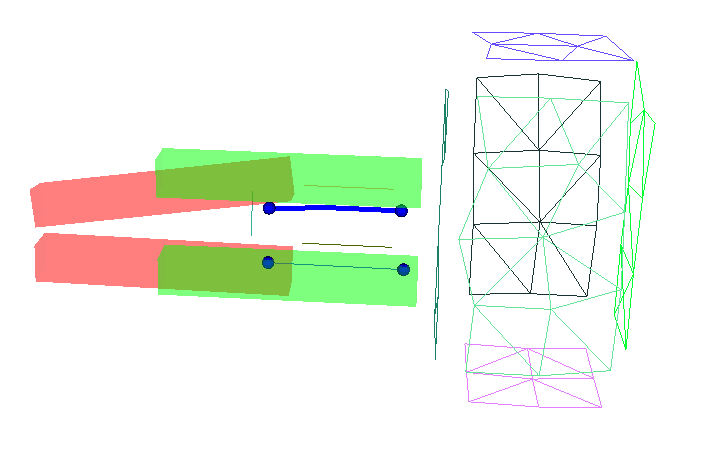}
	\caption{In-hand solution from DMG.}\label{fig_extrinsic_dexterity_graph}
	\end{subfigure}
	\caption{The initial grasp of the left image is shown in red in the right images, and the desired grasp is in green. The image on the bottom right shows the object's DMG and the solution (in blue) to move the fingers to the desired configuration.}
\end{figure}

\begin{figure}
	\centering
	\includegraphics[width=0.5\textwidth]{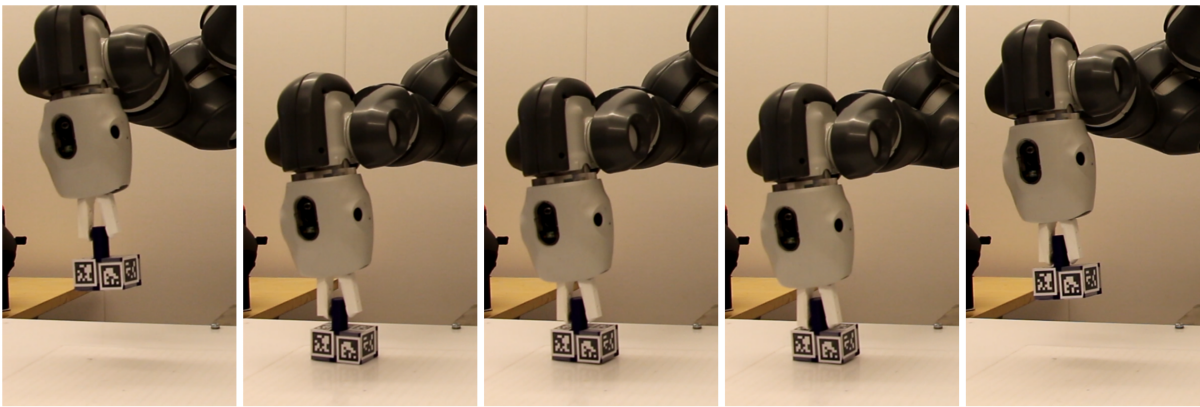}
	\caption{The execution of a translational push using a contact surface instead of the second robot's arm.}\label{fig_extrinsic_dexterity_execution}
\end{figure}

Using the same object, we tested a simple in-hand manipulation task with an external support to help pushing, instead of using the second Yumi's arm. This situation is the one described in section~\ref{sec_single_arm_motion}. We set as contact point for pushing a point on the table in front of Yumi. Therefore, the grasping arm should move with respect to this point to achieve a successful push.
Fig.~\ref{fig_extrinsic_dexterity_initial_grasp} shows the initial grasp of Yumi's gripper on the object. This time, Yumi was grasping the object on the cylindrical part. The desired grasp is shown in Fig.~\ref{fig_extrinsic_dexterity_path}. Fig.~\ref{fig_extrinsic_dexterity_graph} shows the in-hand solution obtained from the DMG.
The solution is composed of an in-hand translation with no rotation along the cylindrical part of the object. Fig.~\ref{fig_extrinsic_dexterity_execution} show the execution of this translation exploiting the table to push the object inside the gripper.
It is important to notice that the pushing strategy that we designed assumes contacts between points, and not surfaces. Therefore, for more complex in-hand tasks, simply replacing the second gripper with a contact surface might lead to undesired behaviors. However, the DMG solution can still be used with a single arm robot, as long as the contacts for pushes are properly designed and, if necessary, taken into account in the graph search phase. The main drawback lies in the requirement to properly model the surrounding environment or to place ad-hoc fixtures to execute the desired pushes.


\section{Conclusions}\label{sec_conclusions}
We have presented and analyzed the Dexterous Manipulation Graph as a tool for object manipulation. We have described how this graph representation is obtained given an object's shape and how to exploit it to plan grasp reconfigurations. More specifically, the DMG can be used both for in-hand manipulation and for more general reconfigurations that involve regrasping. We have formulated the in-hand manipulation execution for a dual-arm robot as a relative motion between the two grippers by using ECTS.
Experiments with Yumi have shown how the application of the DMG helps in planning in-hand manipulation and other manipulation tasks.

As future work, we plan to extend our in-hand manipulation planning to take into account unknown object. This involves a generation of the DMG from the available sensory data, which will only provide an incomplete shape. The DMG should be modified to take into account the lack of full information and have the capability to adapt itself to new available sensory information. In addition, we will explore modifications of the DMG search that are more connected to the dual-arm robot's kinematic structure, so that the robot's arms motion and the object's motion are more dependent on each other and are chosen according to joint optimization criteria. Finally, we plan to adapt the DMG to take into account multi-fingered hands and allow for planning of other dexterous manipulation tasks.

\bibliographystyle{IEEEtran}
\bibliography{General,DexterousManipulation,ExtrinsicDexterity,DualArm}

\begin{thebibliography}{10}
\providecommand{\url}[1]{#1}
\csname url@rmstyle\endcsname
\providecommand{\newblock}{\relax}
\providecommand{\bibinfo}[2]{#2}
\providecommand\BIBentrySTDinterwordspacing{\spaceskip=0pt\relax}
\providecommand\BIBentryALTinterwordstretchfactor{4}
\providecommand\BIBentryALTinterwordspacing{\spaceskip=\fontdimen2\font plus
\BIBentryALTinterwordstretchfactor\fontdimen3\font minus
  \fontdimen4\font\relax}
\providecommand\BIBforeignlanguage[2]{{%
\expandafter\ifx\csname l@#1\endcsname\relax
\typeout{** WARNING: IEEEtran.bst: No hyphenation pattern has been}%
\typeout{** loaded for the language `#1'. Using the pattern for}%
\typeout{** the default language instead.}%
\else
\language=\csname l@#1\endcsname
\fi
#2}}

\bibitem{cruciani_dexterous_manipulation_graph}
S.~Cruciani, C.~Smith, D.~Kragic, and K.~Hang, ``Dexterous manipulation
  graphs,'' in \emph{2018 IEEE/RSJ International Conference on Intelligent
  Robots and Systems (IROS)}, Oct 2018, pp. 2040--2047.

\bibitem{bohg_data_driven_grasp}
J.~{Bohg}, A.~{Morales}, T.~{Asfour}, and D.~{Kragic}, ``Data-driven grasp
  synthesis--a survey,'' \emph{IEEE Transactions on Robotics}, vol.~30, no.~2,
  pp. 289--309, April 2014.

\bibitem{tournassoud_regrasping}
P.~Tournassoud, T.~Lozano-Perez, and E.~Mazer, ``Regrasping,'' in
  \emph{Robotics and Automation. Proceedings. 1987 IEEE International
  Conference on}, vol.~4, Mar 1987, pp. 1924--1928.

\bibitem{furukawa_dynamic_regrasping}
N.~Furukawa, A.~Namiki, S.~Taku, and M.~Ishikawa, ``Dynamic regrasping using a
  high-speed multifingered hand and a high-speed vision system,'' in
  \emph{Proceedings 2006 IEEE International Conference on Robotics and
  Automation, 2006. ICRA 2006.}, May 2006, pp. 181--187.

\bibitem{santina_pisa_soft_hand}
C.~D. Santina, C.~Piazza, G.~Grioli, M.~G. Catalano, and A.~Bicchi, ``Toward
  dexterous manipulation with augmented adaptive synergies: The pisa/iit
  softhand 2,'' \emph{IEEE Transactions on Robotics}, pp. 1--16, 2018.

\bibitem{or_interpolation_control_posture_in-hand}
K.~Or, M.~Tomura, A.~Schmitz, S.~Funabashi, and S.~Sugano, ``Interpolation
  control posture design for in-hand manipulation,'' in \emph{2015 IEEE/SICE
  International Symposium on System Integration (SII)}, Dec 2015, pp. 187--192.

\bibitem{bicchi_dexterous_manipulation_review}
A.~Bicchi, ``Hands for dexterous manipulation and robust grasping: a difficult
  road toward simplicity,'' \emph{IEEE Transactions on Robotics and
  Automation}, vol.~16, no.~6, pp. 652--662, Dec 2000.

\bibitem{okamura_dexterous_manipulation_overview}
A.~M. Okamura, N.~Smaby, and M.~R. Cutkosky, ``An overview of dexterous
  manipulation,'' in \emph{Proceedings 2000 ICRA. Millennium Conference. IEEE
  International Conference on Robotics and Automation. Symposia Proceedings
  (Cat. No.00CH37065)}, vol.~1, 2000, pp. 255--262 vol.1.

\bibitem{ozawa_dexterity_control_survey}
R.~Ozawa and K.~Tahara, ``Grasp and dexterous manipulation of multi-fingered
  robotic hands: a review from a control view point,'' \emph{Advanced
  Robotics}, vol.~31, no. 19-20, pp. 1030--1050, 2017.

\bibitem{bircher_2fingered_gripper_for_reorientation}
W.~G. Bircher, A.~M. Dollar, and N.~Rojas, ``A two-fingered robot gripper with
  large object reorientation range,'' in \emph{2017 IEEE International
  Conference on Robotics and Automation (ICRA)}, May 2017, pp. 3453--3460.

\bibitem{rojas_underactuated_hand_for_in-hand}
N.~Rojas, R.~R. Ma, and A.~M. Dollar, ``The gr2 gripper: An underactuated hand
  for open-loop in-hand planar manipulation,'' \emph{IEEE Transactions on
  Robotics}, vol.~32, no.~3, pp. 763--770, June 2016.

\bibitem{rahman_dexterous_gripper}
N.~Rahman, L.~Carbonari, M.~D'Imperio, C.~Canali, D.~G. Caldwell, and
  F.~Cannella, ``A dexterous gripper for in-hand manipulation,'' in \emph{2016
  IEEE International Conference on Advanced Intelligent Mechatronics (AIM)},
  July 2016, pp. 377--382.

\bibitem{liarokapis_dexterous_adaptive_hands}
M.~V. Liarokapis and A.~M. Dollar, ``Learning task-specific models for
  dexterous, in-hand manipulation with simple, adaptive robot hands,'' in
  \emph{2016 IEEE/RSJ International Conference on Intelligent Robots and
  Systems (IROS)}, Oct 2016, pp. 2534--2541.

\bibitem{chavan-dafle_shape-shifting_gripper}
N.~Chavan-Dafle, K.~Lee, and A.~Rodriguez, ``Pneumatic shape-shifting fingers
  to reorient and grasp,'' \emph{arXiv preprint arXiv:1809.08420}, 2018.

\bibitem{hang_HFTS}
K.~Hang, M.~Li, J.~A. Stork, Y.~Bekiroglu, F.~T. Pokorny, A.~Billard, and
  D.~Kragic, ``Hierarchical fingertip space: A unified framework for grasp
  planning and in-hand grasp adaptation,'' \emph{IEEE Transactions on
  Robotics}, vol.~32, no.~4, pp. 960--972, Aug 2016.

\bibitem{sundaralingam_in-grasp_manipulation}
B.~Sundaralingam and T.~Hermans, ``Relaxed-rigidity constraints: In-grasp
  manipulation using purely kinematic trajectory optimization,'' in
  \emph{Robotics: Science and Systems}, 2017.

\bibitem{psomopoulou_stable_pinching}
E.~Psomopoulou, D.~Karashima, Z.~Doulgeri, and K.~Tahara, ``Stable pinching by
  controlling finger relative orientation of robotic fingers with rolling soft
  tips,'' \emph{Robotica}, vol.~36, no.~2, pp. 204--224, 2018.

\bibitem{saut_dexterous_manipulation_planning_roadmaps}
J.~Saut, A.~Sahbani, S.~El-Khoury, and V.~Perdereau, ``Dexterous manipulation
  planning using probabilistic roadmaps in continuous grasp subspaces,'' in
  \emph{2007 IEEE/RSJ International Conference on Intelligent Robots and
  Systems}, Oct 2007, pp. 2907--2912.

\bibitem{andrychowicz_learning_dexterous_manipulation}
M.~Andrychowicz, B.~Baker, M.~Chociej, R.~Józefowicz, B.~McGrew, J.~Pachocki,
  A.~Petron, M.~Plappert, G.~Powell, A.~Ray, J.~Schneider, S.~Sidor, J.~Tobin,
  P.~Welinder, L.~Weng, and W.~Zaremba, ``Learning dexterous in-hand
  manipulation,'' \emph{arXiv preprint arXiv:1808.00177}, 2017.

\bibitem{chavan-dafle_extrinsic_dexterity}
N.~C. Dafle, A.~Rodriguez, R.~Paolini, B.~Tang, S.~S. Srinivasa, M.~Erdmann,
  M.~T. Mason, I.~Lundberg, H.~Staab, and T.~Fuhlbrigge, ``Extrinsic dexterity:
  In-hand manipulation with external forces,'' in \emph{2014 IEEE International
  Conference on Robotics and Automation (ICRA)}, May 2014, pp. 1578--1585.

\bibitem{sintov_swing-up_regrasping}
A.~Sintov, O.~Tslil, and A.~Shapiro, ``Robotic swing-up regrasping manipulation
  based on the impulse-momentum approach and clqr control,'' \emph{IEEE
  Transactions on Robotics}, vol.~32, no.~5, pp. 1079--1090, Oct 2016.

\bibitem{cruciani_3stages_pivoting}
S.~Cruciani and C.~Smith, ``In-hand manipulation using three-stages open loop
  pivoting,'' in \emph{2017 IEEE/RSJ International Conference on Intelligent
  Robots and Systems (IROS)}, Sept 2017, pp. 1244--1251.

\bibitem{cruciani_integrated_pivoting}
S.~{Cruciani} and C.~{Smith}, ``Integrating path planning and pivoting,'' in
  \emph{2018 IEEE/RSJ International Conference on Intelligent Robots and
  Systems (IROS)}, Oct 2018, pp. 6601--6608.

\bibitem{vina_adaptive_control_pivoting}
F.~E. Vi{\~{n}}a, Y.~Karayiannidis, C.~Smith, and D.~Kragic, ``Adaptive control
  for pivoting with visual and tactile feedback,'' in \emph{2016 IEEE
  International Conference on Robotics and Automation (ICRA)}, May 2016, pp.
  399--406.

\bibitem{antonova_RL_pivoting}
R.~Antonova, S.~Cruciani, C.~Smith, and D.~Kragic, ``Reinforcement learning for
  pivoting task,'' \emph{arXiv preprint arXiv:1703.00472}, 2017.

\bibitem{shi_dynamic_sliding}
J.~Shi, J.~Z. Woodruff, and K.~M. Lynch, ``Dynamic in-hand sliding
  manipulation,'' in \emph{Intelligent Robots and Systems (IROS), 2015 IEEE/RSJ
  International Conference on}, Sept 2015, pp. 870--877.

\bibitem{shi_dynamic_sliding_journal}
J.~Shi, J.~Z. Woodruff, P.~B. Umbanhowar, and K.~M. Lynch, ``Dynamic in-hand
  sliding manipulation,'' \emph{IEEE Transactions on Robotics}, vol.~33, no.~4,
  pp. 778--795, Aug 2017.

\bibitem{eppner_exploitation_environmental_consraints_in_grasping}
C.~Eppner, R.~Deimel, J.~Álvarez Ruiz, M.~Maertens, and O.~Brock,
  ``Exploitation of environmental constraints in human and robotic grasping,''
  \emph{The International Journal of Robotics Research}, vol.~34, no.~7, pp.
  1021--1038, 2015.

\bibitem{hou_fast_planning_for3D_any-pose-reorientation}
Z.~J. Yifan~Hou and M.~T. Mason, ``Fast planning for 3d any-pose-reorienting
  using pivoting,'' in \emph{International Conference on Robotics and
  Automation (ICRA) 2018}.\hskip 1em plus 0.5em minus 0.4em\relax IEEE Robotics
  and Automation Society (RAS), May 2018, pp. 1631--1638.

\bibitem{almeida_dexterous_manipulation_external_contacts}
D.~Almeida and Y.~Karayiannidis, ``Dexterous manipulation with compliant grasps
  and external contacts,'' in \emph{2017 IEEE/RSJ International Conference on
  Intelligent Robots and Systems (IROS)}, Sept 2017, pp. 1913--1920.

\bibitem{chavan-dafle_sampling-based_planner}
N.~Chavan-Dafle and A.~Rodriguez, ``Sampling-based planning of in-hand
  manipulation with external pushes,'' in \emph{International Symposium of
  Robotics Research}, December 2017.

\bibitem{chavan-dafle_in-hand_manipulation_motion_cones}
N.~C. Dafle, R.~Holladay, and A.~Rodriguez, ``In-hand manipulation via motion
  cones,'' in \emph{Proceedings of Robotics: Science and Systems}, Pittsburgh,
  Pennsylvania, June 2018.

\bibitem{smith_dualarm_survey}
C.~Smith, Y.~Karayiannidis, L.~Nalpantidis, X.~Gratal, P.~Qi, D.~V.
  Dimarogonas, and D.~Kragic, ``Dual arm manipulation---a survey,''
  \emph{Robotics and Autonomous Systems}, vol.~60, no.~10, pp. 1340 -- 1353,
  2012.

\bibitem{murooka_whole-body_manipulation}
M.~Murooka, Y.~Inagaki, R.~Ueda, S.~Nozawa, Y.~Kakiuchi, K.~Okada, and
  M.~Inaba, ``Whole-body holding manipulation by humanoid robot based on
  transition graph of object motion and contact,'' in \emph{2015 IEEE/RSJ
  International Conference on Intelligent Robots and Systems (IROS)}, Sept
  2015, pp. 3950--3955.

\bibitem{vahrenkamp_bimanual_grasp_planning}
N.~Vahrenkamp, M.~Przybylski, T.~Asfour, and R.~Dillmann, ``Bimanual grasp
  planning,'' in \emph{2011 11th IEEE-RAS International Conference on Humanoid
  Robots}, Oct 2011, pp. 493--499.

\bibitem{almeida_cooperative_manipulation}
D.~Almeida and Y.~Karayiannidis, ``Cooperative manipulation and identification
  of a 2-dof articulated object by a dual-arm robot,'' in \emph{IEEE
  International Conference on Robotics and Automation (ICRA), 2018}, May 2018.

\bibitem{sommer_bimanual_tactile_eploration}
N.~Sommer, M.~Li, and A.~Billard, ``Bimanual compliant tactile exploration for
  grasping unknown objects,'' in \emph{2014 IEEE International Conference on
  Robotics and Automation (ICRA)}, May 2014, pp. 6400--6407.

\bibitem{xian_closed-chain_dualarm_manipulation}
Z.~Xian, P.~Lertkultanon, and Q.~C. Pham, ``Closed-chain manipulation of large
  objects by multi-arm robotic systems,'' \emph{IEEE Robotics and Automation
  Letters}, vol.~2, no.~4, pp. 1832--1839, Oct 2017.

\bibitem{vahrenkamp_dual_arm_manipulation_regrasping}
N.~Vahrenkamp, D.~Berenson, T.~Asfour, J.~Kuffner, and R.~Dillmann, ``Humanoid
  motion planning for dual-arm manipulation and re-grasping tasks,'' in
  \emph{2009 IEEE/RSJ International Conference on Intelligent Robots and
  Systems}, Oct 2009, pp. 2464--2470.

\bibitem{saut_two_hand_regrasping}
J.~Saut, M.~Gharbi, J.~Cortés, D.~Sidobre, and T.~Siméon, ``Planning
  pick-and-place tasks with two-hand regrasping,'' in \emph{2010 IEEE/RSJ
  International Conference on Intelligent Robots and Systems}, Oct 2010, pp.
  4528--4533.

\bibitem{wan_preparatory_manipulation_planning}
W.~Wan, K.~Harada, and F.~Kanehiro, ``Preparatory manipulation planning using
  automatically determined single and dual arms,'' \emph{arXiv preprint
  arXiv:1812.03274}, 2018.

\bibitem{wan_comparing_singlearm_dualarm_regrasp}
W.~Wan and K.~Harada, ``Developing and comparing single-arm and dual-arm
  regrasp,'' \emph{IEEE Robotics and Automation Letters}, vol.~1, no.~1, pp.
  243--250, Jan 2016.

\bibitem{park_ECTS}
H.~A. Park and C.~S.~G. Lee, ``Extended cooperative task space for manipulation
  tasks of humanoid robots,'' in \emph{2015 IEEE International Conference on
  Robotics and Automation (ICRA)}, May 2015, pp. 6088--6093.

\bibitem{park_ECTS_performance_eval}
------, ``Dual-arm coordinated-motion task specification and performance
  evaluation,'' in \emph{2016 IEEE/RSJ International Conference on Intelligent
  Robots and Systems (IROS)}, Oct 2016, pp. 929--936.

\bibitem{papon_voxel_cloud_connectivity}
J.~Papon, A.~Abramov, M.~Schoeler, and F.~Wörgötter, ``Voxel cloud
  connectivity segmentation - supervoxels for point clouds,'' in \emph{2013
  IEEE Conference on Computer Vision and Pattern Recognition}, June 2013, pp.
  2027--2034.

\bibitem{dijkstra}
E.~W. Dijkstra, ``A note on two problems in connexion with graphs,''
  \emph{Numerische Mathematik}, vol.~1, no.~1, pp. 269--271, Dec 1959.

\bibitem{nguyen_force-closure}
V.~. Nguyen, ``Constructing force-closure grasps,'' in \emph{Proceedings. 1986
  IEEE International Conference on Robotics and Automation}, vol.~3, April
  1986, pp. 1368--1373.

\bibitem{mishra_positive_grips}
B.~Mishra, J.~T. Schwartz, and M.~Sharir, ``On the existence and synthesis of
  multifinger positive grips,'' \emph{Algorithmica}, vol.~2, no.~1, pp.
  541--558, Nov 1987.

\bibitem{chen_antipodal_points}
I.-M. Chen and J.~W. Burdick, ``Finding antipodal point grasps on irregularly
  shaped objects,'' \emph{IEEE Transactions on Robotics and Automation},
  vol.~9, no.~4, pp. 507--512, Aug 1993.

\bibitem{rusu_PCL}
R.~B. Rusu and S.~Cousins, ``{3D is here: Point Cloud Library (PCL)},'' in
  \emph{{IEEE International Conference on Robotics and Automation (ICRA)}},
  Shanghai, China, May 9-13 2011.

\bibitem{olson_apriltag}
E.~Olson, ``{AprilTag}: A robust and flexible visual fiducial system,'' in
  \emph{Proceedings of the {IEEE} International Conference on Robotics and
  Automation ({ICRA})}.\hskip 1em plus 0.5em minus 0.4em\relax IEEE, May 2011,
  pp. 3400--3407.

\bibitem{calli_YCB_dataset}
B.~Calli, A.~Singh, J.~Bruce, A.~Walsman, K.~Konolige, S.~Srinivasa, P.~Abbeel,
  and A.~M. Dollar, ``Yale-cmu-berkeley dataset for robotic manipulation
  research,'' \emph{The International Journal of Robotics Research}, vol.~36,
  no.~3, pp. 261--268, 2017.

\end{thebibliography}

%

\begin{IEEEbiography}[{\includegraphics[width=1in,height=1.25in,clip,keepaspectratio]{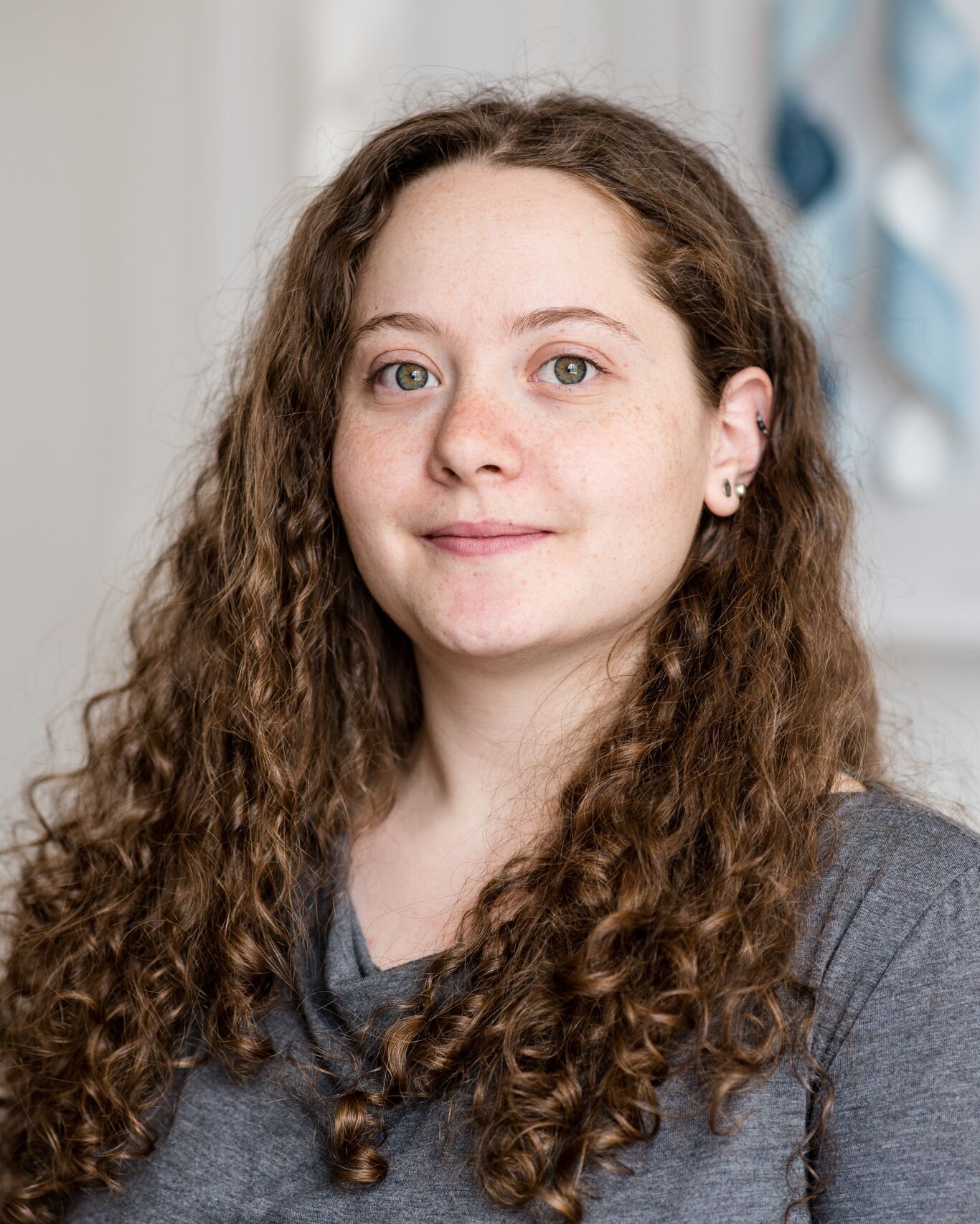}}]
	{Silvia Cruciani}
received the B.S. degree in Computer Science and Control Engineering in 2013 and the M.S. degree in Artificial Intelligence and Robotics in 2015 from Sapienza University of Rome, Italy. She is currently a Ph.D. student at the Robotics, Perception and Learning Lab at KTH Royal Institute of Technology, Stockholm, Sweden. Her research is focused on vision-based dexterous manipulation and extrinsic dexterity.
\end{IEEEbiography}

\begin{IEEEbiography}[{\includegraphics[width=1in,height=1.25in,clip,keepaspectratio]{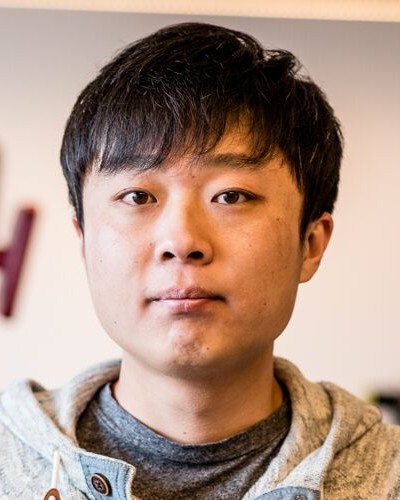}}]{Kaiyu Hang}
	is a postdoctoral associate with the Department of Mechanical Engineering and Material Science, Yale University, New Haven, USA. He received his Ph.D. in Computer Science, specialized in Robotics and Computer Vision in 2016 from KTH Royal Institute of Technology, Stockholm, Sweden. Before that, he received the B.S. degree in information and communication engineering from Xi’an Jiaotong University, Xi’an, China in 2010 and the M.Sc. degree in communication systems from KTH Royal Institute of Technology, Stockholm, Sweden in 2012. His research interests include dexterours grasping and manipulation, mobile manipulation, and robot end-effector design and optimization.
\end{IEEEbiography}

\begin{IEEEbiography}[{\includegraphics[width=1in,height=1.25in,clip,keepaspectratio]{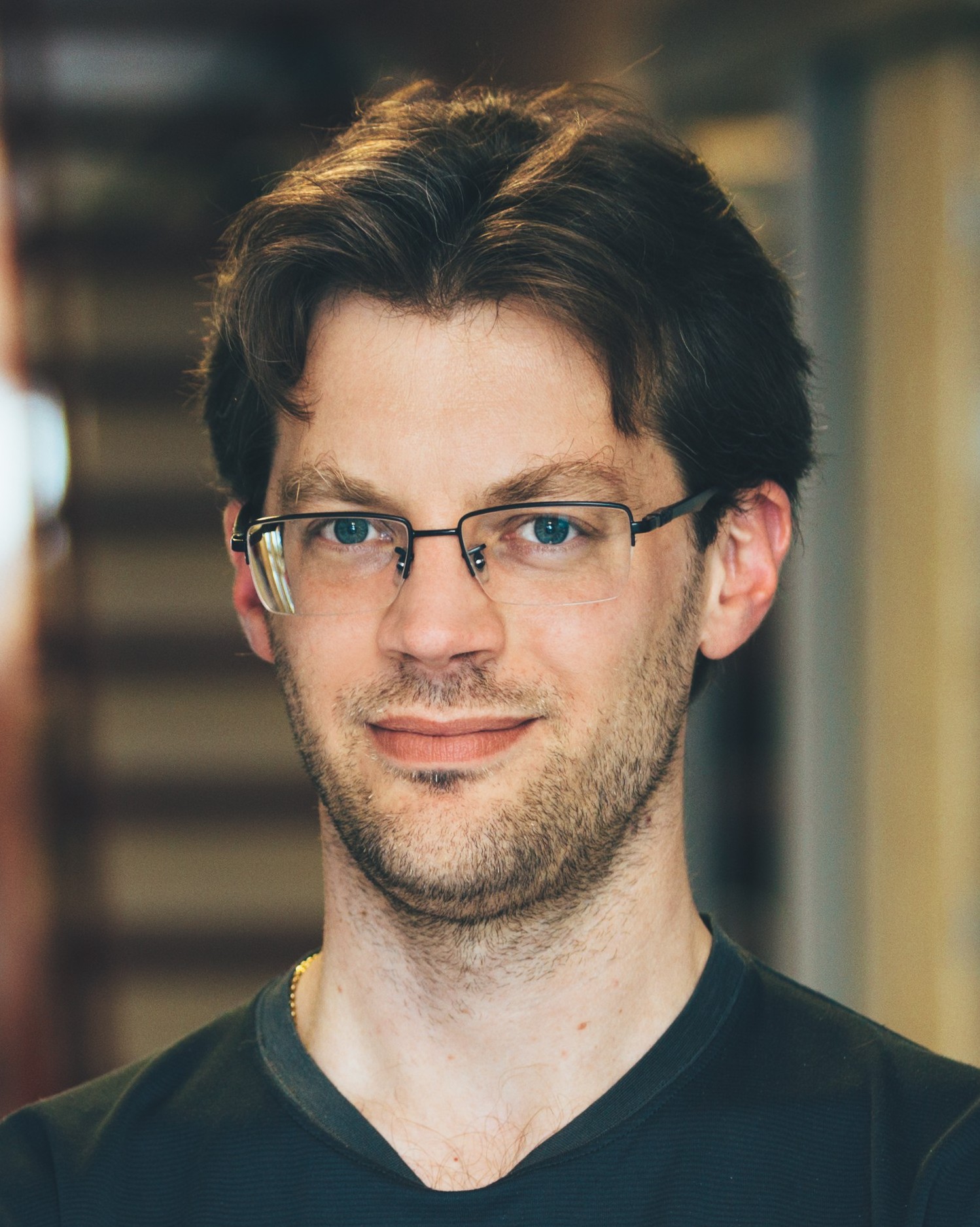}}]{Christian Smith}
is an Associate Professor in Computer Science at the Division of Robotics, Perception, and Learning at The Royal Institute of Technology in Stockholm (KTH). He received his M.Sc. in Engineering Physics in 2005, and Ph.D. in Computer Science in 2009, both at KTH. Between 2010-2011, he was a post-doctoral researcher at Advanced Telecommunications Research International (ATR) in Kyoto, Japan. He is the secretary of the IEEE Robotics and Automation Society Swedish Chapter. Research interests include control and modelling for manipulation and grasping in human-centric environments and human–robot interaction.
\end{IEEEbiography}


\begin{IEEEbiography}[{\includegraphics[width=1in,height=1.25in,clip,keepaspectratio]{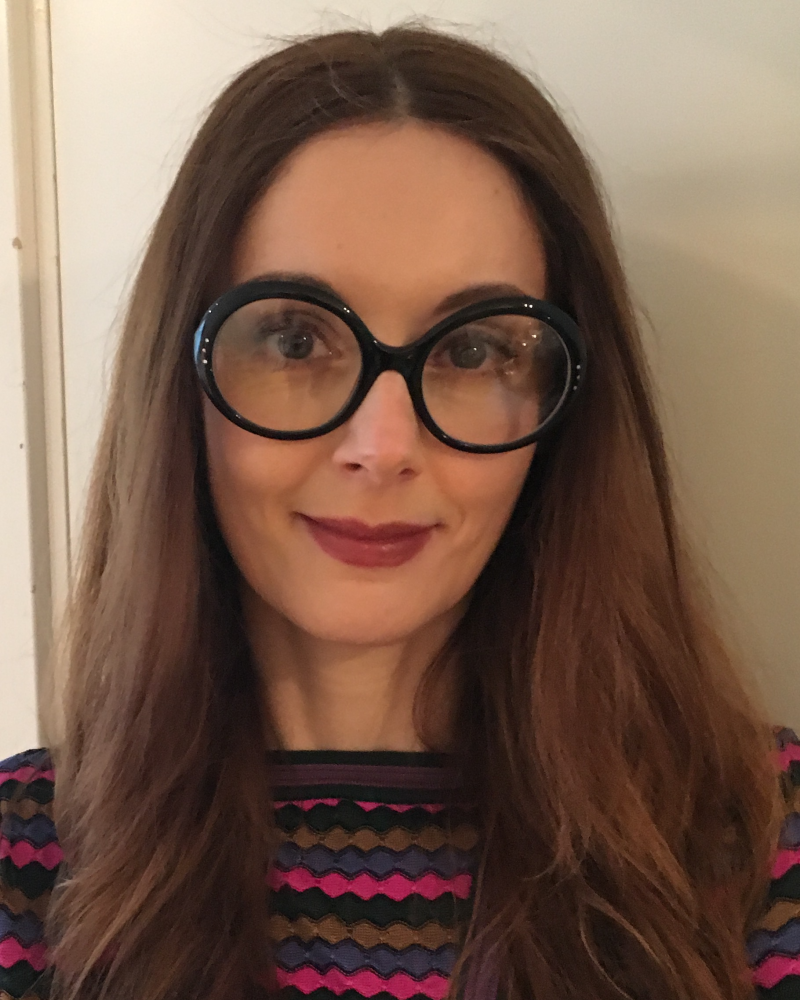}}]{Danica Kragic}
 is a Professor at the School of Electrical Engineering and Computer Science at the Royal Institute of Technology, KTH. She received MSc in Mechanical Engineering from the Technical University of Rijeka, Croatia in 1995 and PhD in Computer Science from KTH in 2001. She has been a visiting researcher at Columbia University, Johns Hopkins University and INRIA Rennes. She is the Director of the Centre for Autonomous Systems. Danica received the 2007 IEEE Robotics and Automation Society Early Academic Career Award. She is a member of the Royal Swedish Academy of Sciences, Royal Swedish Academy of Engineering Sciences and Young Academy of Sweden. She holds a Honorary Doctorate from the Lappeenranta University of Technology. She chaired IEEE RAS Technical Committee on Computer and Robot Vision and served as an IEEE RAS AdCom member. Her research is in the area of robotics, computer vision and machine learning. In 2012, she received an ERC Starting Grant. Her research is supported by the EU, Knut and Alice Wallenberg Foundation, Swedish Foundation for Strategic Research and Swedish Research Council. She is an IEEE Fellow. 
\end{IEEEbiography}




\end{document}